\documentclass{article}

\usepackage{hyperref}

\usepackage{fullpage}
\usepackage[authoryear,sort]{natbib}
\usepackage[utf8]{inputenc} \usepackage[T1]{fontenc}  \usepackage{etoolbox}
\usepackage{microtype}    
\usepackage{comment}
\usepackage{abstract}     
\usepackage{amsmath,amssymb,mathtools,amsfonts}
\usepackage{amsthm}
\usepackage{nicefrac}       
\usepackage{enumitem}

\usepackage{cleveref}
\usepackage{url}            
\usepackage{listings}
\usepackage{multirow}
\usepackage{stackengine} 

\usepackage{algorithm}
\usepackage{algorithmic}
\usepackage{graphicx}
\usepackage{booktabs}
\usepackage{mathrsfs}

\usepackage{color}

\usepackage{subfigure}

\allowdisplaybreaks[4]

\newcounter{ass_counter}
\newcounter{thm_counter}
\newcounter{remark_counter}
\newtheorem{theorem}[thm_counter]{Theorem}\newtheorem{lemma}[thm_counter]{Lemma}\newtheorem{corollary}[thm_counter]{Corollary}
\newtheorem{assumption}[ass_counter]{Assumption}
\newtheorem{remark}[remark_counter]{Remark}
\Crefname{assumption}{Assumption}{Assumptions}
\newtheorem{definition}{Definition}
\usepackage{antpolt}
\usepackage{caption}

\title{Faster Distributed Deep Net Training: Computation and Communication Decoupled
 Stochastic Gradient Descent}

\date{\today}

\usepackage{authblk}
 \author[1,2]{Shuheng Shen}
 \author[1,2]{Linli Xu\footnote{The corresponding author.}}
 \author[1,2]{Jingchang Liu}
 \author[1,2]{Xianfeng Liang}
 \author[1,3]{Yifei Cheng}

 \affil[ ]{\texttt{vaip@mail.ustc.edu.cn, linlixu@ustc.edu.cn,  \{xdjcl, zeroxf, chengyif\}@mail.ustc.edu.cn }}
 \affil[1]{Anhui Province Key Laboratory of Big Data Analysis and Application}
 \affil[2]{School of Computer Science and Technology, University of Science and Technology of China}
 \affil[3]{School of Data Science, University of Science and Technology of China}

\begin{document}

\maketitle

\begin{abstract}

With the increase in the amount of data and the expansion of model scale, distributed parallel training becomes an important and successful technique to address the optimization challenges. Nevertheless, although distributed stochastic gradient descent (SGD) algorithms can achieve a linear iteration speedup, they are limited significantly in practice by the communication cost, making it difficult to achieve a linear time speedup. In this paper, we propose a computation and communication decoupled stochastic gradient descent (CoCoD-SGD) algorithm to run computation and communication in parallel to reduce the communication cost. We prove that CoCoD-SGD has a linear iteration speedup with respect to the total computation capability of the hardware resources. In addition, it has a lower communication complexity and better time speedup comparing with traditional distributed SGD algorithms. Experiments on deep neural network training demonstrate the significant improvements of CoCoD-SGD: when training ResNet18 and VGG16 with 16 Geforce GTX 1080Ti GPUs, CoCoD-SGD is up to 2-3$\times$ faster than traditional synchronous SGD.

 \end{abstract}

\section{Introduction}
\label{sec:introduction}

The training of deep neural networks is resource intensive and time-consuming. 
With the expansion of data and model scale, it may take a few days or weeks to train a deep model by using mini-batch SGD 
on a single machine/GPU. To accelerate the training process, distributed optimization provides an effective tool for deep net training by allocating the computation to multiple computing devices (CPUs or GPUs).

When variants of mini-batch SGD are applied to a distributed system, communication between computing devices will be incurred to keep the same convergence rate as mini-batch SGD. 
As a matter of fact, the extra communication cost in a distributed system is the main factor which prevents a distributed optimization algorithm from achieving the linear time speedup, although the computation load is the same as its single machine version. 
In addition, the communication cost, which is often linearly proportional to the number of workers, can be extremely expensive when the number of workers is huge.
Therefore, it is critical to reduce the communication bottleneck to make better use of the hardware resources.

Given that the total amount of communication bits equals the number of communications multiplied by the number of bits per communication, several works are proposed to accelerate training by reducing the communication frequency~\citep{zhou2017convergence,yu2018parallel,stich2018local} or communication bits~\citep{lin2017deep,stich2018sparsified,wen2017terngrad,alistarh2017qsgd}.
However, even when the communication frequency or the number of bits per communication is reduced, hardware resources are not fully exploited in traditional synchronous distributed algorithms because of the following two reasons: (1) only partial resources can be used when workers are communicating with each other and (2) the computation and the communication are interdependent in each iteration.
Specifically, the computation of the $t$-th step relies on the communication of the $(t-1)$-th step and the communication in the $t$-th step follows the computation of the $t$-th step. 
To address that, another line of research aims to run computation and communication in parallel by using stale gradients~\citep{li2018pipe}, but the communication complexity is still high as all gradients need to be communicated.

To tackle this dilemma, we propose computation and communication decoupled stochastic gradient descent (CoCoD-SGD) for distributed training.
Instead of waiting for the completion of communication as in traditional synchronous SGD, workers in CoCoD-SGD continue to calculate stochastic gradients and update models locally after the communication started.
After the communication finishes, each worker updates its local model with the result of communication and 
the difference of the local models.
As a result, CoCoD-SGD can make full use of resources by running computation and communication in parallel.
In the meantime, workers in CoCoD-SGD communicate with each other periodically instead of in each iteration.
As a benefit, the communication complexity of CoCoD-SGD is lower. By running computation and communication in parallel while reducing the communication complexity, faster distributed training is achieved.
In addition, CoCoD-SGD is suitable for both homogeneous and heterogeneous environments. 
Further, it is theoretically justified with a linear iteration speedup with respect to the total computation capability of hardware devices.

Contributions of this paper are summarized as follows:
\begin{itemize}
  \setlength{\itemsep}{0pt}
  \setlength{\parsep}{0pt}
  \setlength{\parskip}{0pt}
\item We propose CoCoD-SGD, a computation and communication decoupled distributed algorithm, to make full use of hardware resources and reduce the communication complexity.
\item From the theoretical perspective, we prove that CoCoD-SGD has a linear iteration speedup with respect to the number of workers in a homogeneous environment. Besides, it has a theoretically justified linear iteration speedup with respect to the total computation capability in a heterogeneous environment.
\item When training deep neural networks with multiple GPUs, CoCoD-SGD achieves a better time speedup comparing with existing distributed algorithms.
\item In both homogeneous and heterogeneous environments, experimental results verify the effectiveness of the proposed algorithm.
\end{itemize}

\section{Related work}
\label{sec:related-work}

A conventional framework for distributed training is the centralized parameter server architecture~\citep{li2014communication}, which is supported by most existing systems such as Tensorflow~\citep{abadi2016tensorflow}, Pytorch~\citep{paszke2017automatic} and Mxnet~\citep{chen2015mxnet}. 
In each iteration, the parameter server, which holds the global model, needs to communicate with $O(N)$ workers. This becomes a bottleneck which slows down the convergence when $N$ is large. 
Therefore, the decentralized ring architecture using the Ring-AllReduce algorithm
became popular in recent years. In Ring-AllReduce, each worker only transports $O(1)$ gradients to its neighbors to get the average model over all workers.

The Ring-AllReduce algorithm can be directly combined with synchronous stochastic gradient descent (S-SGD), which has a theoretically justified convergence rate $O(\frac{1}{\sqrt{TM}})$ for both general convex~\citep{dekel2012optimal} and non-convex problems~\citep{ghadimi2013stochastic}, where $T$ is the number of iterations and $M$ is the mini-batch size. 
Such rate shows its linear iteration speedup with respect to the number of workers since increasing the number of workers is equivalent to increasing the mini-batch size. 
Although only $O(1)$ gradients are needed to be communicated per worker in each iteration for Ring-AllReduce, $O(N)$ handshakes are needed for each worker. 
Therefore, the communication cost grows as the number of the workers increases.

There have been many attempts to reduce the communication bottleneck of S-SGD while maintaining its linear iteration speedup property. Among them,
Local-SGD is a variant of S-SGD with low communication frequency, in which workers update their models locally and communicate with each other every $k$ iterations. 
It has been proved to have linear iteration speedup for both strongly convex~\citep{stich2018local} and non-convex~\citep{zhou2017convergence,yu2018parallel} problems. 
QSGD~\citep{alistarh2017qsgd} and TernGrad~\citep{wen2017terngrad} compress the gradients 
from 32-bit float to lower bit representations. 
Sparse SGD~\citep{aji2017sparse} is another method that communicates part of the gradients in each iteration and is proved to have the same convergence rate as SGD~\citep{stich2018sparsified,alistarh2018convergence}. 
Although the communication complexity is reduced in the above methods, the hardware resources are not fully utilized because only a part of them is used for communication.

To fully utilize hardware resources, asynchronous stochastic gradient descent (A-SGD), a distributed variant of SGD based on the parameter server architecture, is proposed. 
After one communication started, a worker uses a stale model to compute the next stochastic gradient without waiting for the completion of the communication.
A linear iteration speedup is proved for both convex~\citep{agarwal2011distributed} and non-convex~\citep{lian2015asynchronous} problems.
However, stale gradients may slow down the training and make it converge to a poor solution especially when the number of workers is large, which implies a large delay~\citep{chen2016revisiting}. 
As a remedy, Pipe-SGD~\citep{li2018pipe} is proposed to integrate the advantages of A-SGD and Ring-AllReduce. Specifically, it employs Ring-AllReduce to get the average gradient and updates the model with a stale gradient. 
Pipe-SGD can control the delay as a constant because of the efficiency of the Ring-AllReduce algorithm. 
Nevertheless, one disadvantage of Pipe-SGD is that its communication complexity is $O(T)$, which is higher than Local-SGD and QSGD.
In comparison, the proposed algorithm which makes full use of hardware resources and has 
a lower communication complexity, achieves the advantages of both Pipe-SGD and Local-SGD.

\section{Algorithm}

In this section, we introduce the CoCoD-SGD algorithm with the following three techniques: (1) computation and communication decoupling, (2) periodically communicating and (3) proportionally sampling.

\subsection{Preliminary}
\subsubsection{Problem Definition}
We focus on data-parallel distributed training, in which 
each worker can access only part of the data.
We use $\mathcal{D}$ to denote the full training dataset and $\mathcal{D}_i$ to denote the local dataset stored in the $i$-th worker. 
We have $\mathcal{D} = \mathcal{D}_1 \cup \cdots \cup \mathcal{D}_N$ and $\mathcal{D}_i \cap \mathcal{D}_j = \varnothing, \forall i \neq j$.
The objective can be written as 
\begin{equation}\label{basic_object}
\min_{x \in \mathbb{R}^d} f(x) := \frac{1}{|\mathcal{D}|} \sum_{\xi \in \mathcal{D}} f(x, \xi) = \sum_{i=1}^{N} p_i f_i(x),
\end{equation}
where $f_i(x) := \frac{1}{|\mathcal{D}_i|} \sum_{\xi_i \in \mathcal{D}_i} f(x, \xi_i)$ is the local loss of the $i$-th worker and $p_i$'s define the partition of data among all workers. 
Specifically, $p_i$ is proportional to size of the local dataset on the $i$-th worker: $p_i = \frac{|\mathcal{D}_i|}{|\mathcal{D}|}$ and we have $\sum_{i=1}^{N} p_i = 1$. 
We use $\mathcal{P}$ to denote the distribution of $p_i$'s.

\subsubsection{Notations}
\begin{itemize}
\item $\| \cdot \|$ indicates the $\ell_2$ norm of a vector.
\item $f^*$ represents the optimal value of (\ref{basic_object}).
\item $\mathbb{E}$ indicates a full expectation with respect to all the randomness, which includes the random indexes sampled to calculate stochastic gradients in all iterations.
\item $\mathcal{C}_i$ represents the computation capability, which indicates the computing speed, of the $i$-th worker.
\end{itemize}

\subsection{Computation and Communication Decoupled Stochastic Gradient Descent}
The complete procedure of CoCoD-SGD is summarized in Algorithm~\ref{CoCoD-SGD}. 
On one hand, the goal of  CoCoD-SGD is two-fold: (1) running computation and communication in parallel and (2) reducing the communication complexity. To achieve that, we propose the following two techniques:
\begin{itemize}

\item \textbf{Computation and communication decoupling:} 
Different from Pipe-SGD which uses stale gradients to decouple the dependency of computation and communication, CoCoD-SGD continues to update models in workers locally after the communication starts. 
If one communication starts after the $t$-th update, workers need to communicate with each other to get the average model $\hat{x}_t$.
After the communication starts, the $i$-th worker continues to update its local model $x_{t}^i$ by mini-batch SGD (line 4-5). 
As a result, computation and communication can be executed simultaneously.

\item \textbf{Periodically communicating:} In practice, when the number of workers is large, the communication time may exceed the time of one update, which will cause idle time of the computing devices. Therefore, we let workers keep on updating the local models $k$ times instead of waiting for the completion of the communication after only one local update (line 5). In this way, the communication complexity can be also reduced.
\end{itemize}

\begin{algorithm}[!tb]
\caption{CoCoD-SGD}
\label{CoCoD-SGD}
\hspace*{0.02in}{\bf Input:}
The number of workers $N$, the number of iterations $T$, communication period $k$, mini-batch sizes $M_1,\cdots, M_N$ and initial point $\hat{x}_{0} \in \mathbb{R}^{d}$. 

\hspace*{0.02in}{\bf Initialize:}
$t=0, x_{i}^{0} = \hat{x}_0, i=1, 2, \cdots, N$.

\begin{algorithmic}[1]
\WHILE{$t < T$}
  \STATE \underline{Worker $W_i$ does}: 
  \STATE Run \textbf{Step \uppercase\expandafter{\romannumeral1} } and \textbf{Step \uppercase\expandafter{\romannumeral2} } in parallel.
  \STATE \textbf{Step \uppercase\expandafter{\romannumeral1} }: Store $x_t^i$ in the memory and communicate with other workers to get the weighted mean of all local models: $\hat{x}_t = \sum_{i=1}^N \frac{M_i}{\sum_{j=1}^{N} M_j} x_t^i$.
  \STATE \textbf{Step \uppercase\expandafter{\romannumeral2} }: 
  \begin{enumerate}
    \setlength{\itemindent}{0.5em}
    \item[] \textbf{for} $\tau = t$ to $t+k-1$ \textbf{do}
    \item[] \ \ \ \ \ Compute a mini-batch stochastic gradient $G_{\tau}^i = \frac{1}{M_i} \sum_{j=1}^{M_i} \nabla f_i(x_{\tau}^i, \xi_{\tau}^{i,j}), \xi_{\tau}^{i,j} \in \mathcal{D}_i$, and update
    \item[] \  locally: $x_{\tau + 1}^i = x_{\tau}^i - \gamma G_{\tau}^i$
    \item[] \textbf{end for}
  \end{enumerate} 
  \STATE $x_{t+k}^i = \hat{x}_t + (x_{t+k}^i  - x_t^i)$
  \STATE $t = t+k $
\ENDWHILE
\end{algorithmic}
\end{algorithm}

On the other hand, CoCoD-SGD is designed to be suitable for heterogeneous distributed system, where workers may have different computation capabilities. 
Intuitively, when all workers use the same batch size to compute stochastic gradients, the faster workers need to wait for the slower ones after finishing their updates. 
Therefore, we use proportionally sampling in CoCoD-SGD:
\begin{itemize}
  \item \textbf{Proportionally sampling:} Workers in CoCoD-SGD use different mini-batch sizes $M_i$'s to compute stochastic gradients.
   The batch sizes are proportional to the computation capabilities of workers, i.e., $\frac{M_i}{M_j} = \frac{\mathcal{C}_i}{\mathcal{C}_j}$. 
   Under this setting, all workers will finish $k$ updates at the same time. 
   Meanwhile, to let all workers finish one epoch simultaneously, we proportionally divide the dataset among workers, i.e., $\frac{|\mathcal{D}_i|}{|\mathcal{D}_j|} = \frac{\mathcal{C}_i}{\mathcal{C}_j}$. 
   In addition, we define the average model $\hat{x}_t$ as the weighted mean of all local models: $\hat{x}_t = \sum_{i=1}^N \frac{M_i}{\sum_{j=1}^{N} M_j} x_t^i$ (line 4).
\end{itemize}

Besides, different from QSGD~\citep{alistarh2017qsgd} and Sparse-SGD~\citep{stich2018sparsified}, which are based on the parameter server architecture, CoCoD-SGD can communicate with the Ring-AllReduce algorithm by the definition of $\hat{x}^t$.

After finishing local iterations and the communication, the $i$-th worker updates its local model with the average model and 
the difference of the local models
by $x_{t+k}^i = \hat{x}_t + (x_{t+k}^i  - x_t^i)$ (line 6).

\section{Theoretical Analysi}

In this section, we provide the theoretical analysis for CoCoD-SGD and show that CoCoD-SGD has the same convergence rate as S-SGD. In addition, we show that CoCoD-SGD has lower communication complexity and better time speedup.
Due to the space limit, all proofs are deferred to the supplemental material.
In the subsequence analysis, we will use the following definitions.

\begin{definition} We denote the total number of iterations and the total time used to converge when using $N$ workers as $T_N$ and $\mathcal{T}_N$, respectively. Then the 
iteration speedup (IS) and the time 
speedup (TS) are respectively defined as
  \begin{equation}\label{def_is}
  {\rm IS}^N = \frac{T_1}{T_N},
  \end{equation}
  \begin{equation}\label{def_ts}
  {\rm TS}^N = \frac{\mathcal{T}_1}{\mathcal{T}_N}.
  \end{equation}
\end{definition}

\subsection{Main Results}
Before establishing our main results, we introduce the following assumptions, all of which are commonly used in the analysis of distributed algorithms~\citep{lian2015asynchronous,aji2017sparse,yu2018parallel}.
\begin{assumption} \label{assumptions} ~~
\begin{itemize}
\item[(1)] \textbf{Lipschitz gradient}: All local functions $f_{i}$'s have $L$-Lipschitz gradients
      \begin{equation}
      \| \nabla f_i(x) - \nabla f_i(y) \| \leq L \| x - y \|, \forall i , \forall x, y \in \mathbb{R}^d.
      \end{equation}
\item[(2)] \textbf{Unbiased estimation}: 
      \begin{eqnarray}
      \mathbb{E}_{\xi \sim \mathcal{D}_i} \nabla f_{i} (x, \xi) &=& \nabla f_{i}(x), \forall i,
      \\
      \mathbb{E}_{i \sim \mathcal{P}} \mathbb{E}_{\xi \sim \mathcal{D}_i} \nabla f_{i} (x, \xi) &=& \nabla f(x).
      \end{eqnarray}
\item[(3)] \textbf{Bounded variance}: There exist constants $\sigma$ and $\zeta$ such that
      \begin{eqnarray}
      \mathbb{E}_{\xi \sim \mathcal{D}_i} \| \nabla f_{i} (x, \xi) - \nabla f_{i} (x)\|^2 \leq \sigma^2, ~~\forall x \in \mathbb{R}^d, \forall i,
      \\
      \mathbb{E}_{i \sim \mathcal{P}} \| \nabla f_i(x) - \nabla f(x) \|^2 \leq \zeta^2, ~~\forall x \in \mathbb{R}^d.~~~~~~
      \end{eqnarray}
\item[(4)] \textbf{Dependence of random variables}: $\xi_{t}^{i,j}$'s are independent random variables, where $t \in \{0, 1, \cdots, T-1\}$, $i \in \{1,2, \cdots, N\}$, and $j \in \{1,2, \cdots, M_i\}$. 
\end{itemize}
\end{assumption}

To evaluate the convergence rate, the metric in nonconvex optimization is to bound the weighted average of the  $\ell_2$ norm of all gradients~\citep{ghadimi2013stochastic,lian2015asynchronous,yu2018parallel}.
\begin{theorem} \label{general_theorem}
Under Assumption~\ref{assumptions}, if the learning rate satisfies $\gamma \leq \frac{1}{L}$, we have the following convergence result for Algorithm~\ref{CoCoD-SGD}:
{\small
\begin{eqnarray} \label{general_threorem_result}
\frac{1}{T} \sum_{t=0}^{T-1} D_1 \mathbb{E} \| \nabla f(\hat{x}_t) \|^2
\leq\frac{ 2(f(\hat{x}_0) - f^*)}{T\gamma} +  D_2 (\frac{N\sigma^2}{\sum_{i=1}^{N} M_i} + 2k\zeta^2) + \frac{ \gamma L \sigma^2}{\sum_{i=1}^{N} M_i},
\end{eqnarray}
}
where $k$ is the communication period and
\begin{equation}
D_1 = 1 - 2kD_2, ~~~~D_2 = \frac{8 \gamma^2 L^2 k}{1 - 16 \gamma^2 k^2 L^2}.
\end{equation}
\end{theorem}

Choosing the learning rate $\gamma$ appropriately, 
we have the following corollary.
\begin{corollary} \label{general_corollary}
Under Assumption~\ref{assumptions},
when the learning rate is set as $\gamma = \frac{1}{\sigma \sqrt{\frac{T}{\sum_{i=1}^{N} M_i}}}$ and the total number of iterations satisfies
{\small
\begin{eqnarray}\label{convergence_condition}
T \geq \max \Bigg \{ \frac{L^2(\sum_{i=1}^{N} M_i)}{\sigma^2}, \frac{48(\sum_{i=1}^{N} M_i)L^2k^2}{\sigma^2},  
\frac{144(\sum_{i=1}^{N} M_i)^3}{\sigma^6}L^2k^2 \left( \frac{N\sigma^2}{\sum_{i=1}^{N} M_i} + 2k\zeta^2 \right)^2 \Bigg\},
\end{eqnarray}
}
we have the following convergence result for Algorithm~\ref{CoCoD-SGD}:
\begin{equation}\label{convergence_rate}
\frac{1}{T} \sum_{t=0}^{T-1} \mathbb{E} \| \nabla f(\hat{x}_t) \|^2 \leq \frac{4\sigma(f(\hat{x}_0) - f^* + L)}{\sqrt{T\sum_{i=1}^{N} M_i}}.
\end{equation}
\end{corollary}

Corollary~\ref{general_corollary} indicates that the convergence rate of the weighted average model is $O\left(1/\sqrt{\sum_{i=1}^{N} M_i T} \right)$, which is consistent with S-SGD~\citep{ghadimi2013stochastic}. 
Next, we establish the linear iteration speedup in both homogeneous and heterogeneous environments, and show the communication complexity of CoCoD-SGD.

\begin{remark}
 {\rm(linear iteration speedup in the homogeneous environment)}. For CoCoD-SGD in a homogeneous environment, all workers use the same mini-batch size: $M_1 = M_2 = \cdots = M_N = M$. According to (\ref{convergence_rate}), CoCoD-SGD converges at the rate $O\left(1/\sqrt{NMT}\right)$. Consequently, to achieve the $\epsilon$-approximation solution, $O(1/ (NM\epsilon^2)$ iterations are needed, which means CoCoD-SGD has a linear iteration speedup with respect to the number of workers according to the definition of IS in (\ref{def_is}).
\end{remark}

\begin{remark}
{\rm (linear iteration speedup in the heterogeneous environment)}.
With Proportionally Sampling, we have $M_i / M_j = \mathcal{C}_i / \mathcal{C}_j$. 
According to (\ref{convergence_rate}), to achieve the $\epsilon$-approximation solution, the number of iterations required is $O\left(1/(\sum_{i=1}^{N} M_i \cdot \epsilon^2)\right)$, that is $O\left( \mathcal{C}_1 / \left(\sum_{i=1}^{N} \mathcal{C}_i \cdot M_1 \epsilon^2 \right) \right)$, which means CoCoD-SGD has a linear iteration speedup with respect to the total computation capability of all workers according to (\ref{def_is}).
\end{remark}

\begin{remark} \label{remark_3}
{\rm (communication complexity)}. 
From (\ref{convergence_condition}), we have $T \geq \max \{ O(N), O(N k^2), O(N^3 k^4 ) \}$. Therefore, when the communication period is bounded by  $O\left(T^{\frac{1}{4}}/N^{\frac{3}{4}} \right)$, the convergence rate in Corollary~\ref{general_corollary} is achievable. 
As a result, the total communication complexity of CoCoD-SGD is $O\left(T/\left(T^{\frac{1}{4}}/N^{\frac{3}{4}} \right)\right)$, that is $O\left(T^{\frac{3}{4}} N^{\frac{3}{4}}\right)$. 
\end{remark}

\begin{remark}\label{remark_4}
{\rm (choice of learning rate)}. When we run CoCoD-SGD for a fixed number of epochs, $\sum_{i=1}^N M_i T$ will be a constant. The learning rate suggested in Corollary~\ref{general_corollary} can be written as $\gamma = \frac{\sum_{i=1}^N M_i}{\sigma \sqrt{\sum_{i=1}^N M_i T}}$. Thus, for CoCoD-SGD with $N$ workers, the learning rate should be set as $\gamma_N = N \cdot \gamma_1$ in the homogeneous environment and $\gamma_N = \sum_{i=1}^{N} \mathcal{C}_i / \mathcal{C}_1 \cdot \gamma_1$ in the heterogeneous environment.
\end{remark}

\begin{figure}[t]
\centering
\includegraphics[scale=0.36]{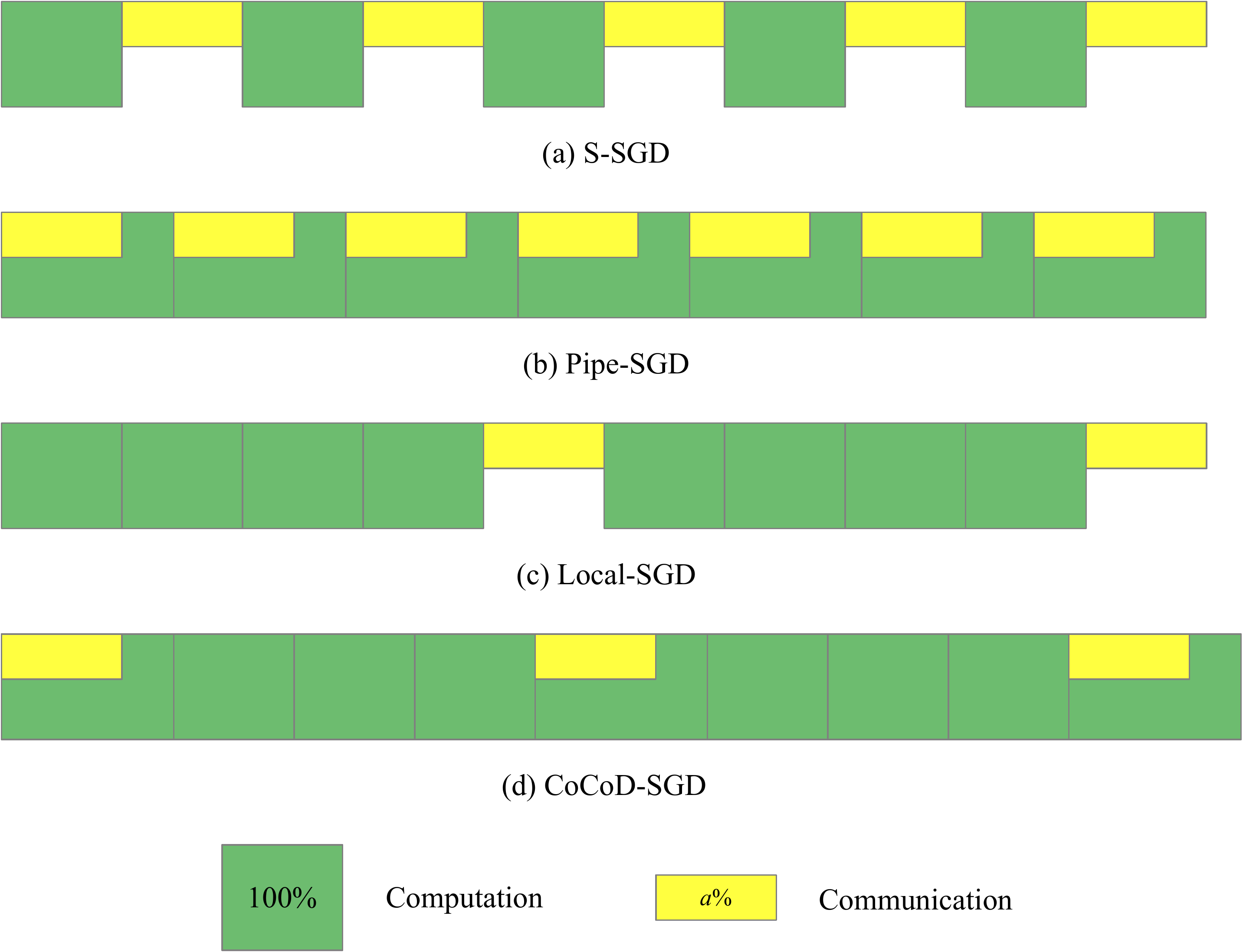}
\caption{Comparison of S-SGD, Pipe-SGD, Local-SGD and CoCoD-SGD. A green block indicates computation which can make full use of the hardware resources, and a yellow block corresponds to communication which only uses $a\%$ of the resources. S-SGD and Local-SGD run computation and communication sequentially, while Pipe-SGD and CoCoD-SGD run them in parallel. Both Local-SGD and CoCoD-SGD communicate every $k$ iterations.}
\label{fig:baseline_comparison}
\end{figure}

\subsection{Time Speedup Analysis}
Next, we compare and analyse the time speedup of CoCoD-SGD and other distributed algorithms when they are applied in practice.

When multiple nodes with GPUs are used to train a deep neural network, 
GPUDirect communication, 
where all computation threads and communication threads are executed in GPUs, is widely adopted~\citep{paszke2017automatic,chen2015mxnet,sergeev2018horovod} and verified to be efficient~\citep{potluri2013efficient}. 
The computation threads can make full use of GPU resources especially when the size of mini-batch is large, while the communication threads can only use partial of the GPU resources, which is assumed to be $a\%$ for ease of analysis.
We further denote the time of one computation as $\mathcal{T}_{\rm comp}$ and the time of communication when using $N$ workers as $\mathcal{T}_{\rm comm}^N$.

A comparison of the running processes of CoCoD-SGD and other standard distributed algorithms is shown in Figure~\ref{fig:baseline_comparison}. 
All algorithms have provable linear iteration speedup. 
Thus, the time speedup equals $(N \cdot \mathcal{T}_1^{\rm av}) / \mathcal{T}_N^{\rm av}$ according to (\ref{def_ts}), where $\mathcal{T}_N^{\rm av}$ is denoted as the average time required to finish one iteration among $N$ workers.
For S-SGD, computation and communication are sequential.
Therefore, its time speedup is 
\begin{equation}\label{TS_S-SGD}
{\rm TS}_{\rm S\textrm{-}SGD}^N = \frac{N \cdot \mathcal{T}_{\rm comp}}{\mathcal{T}_{\rm comp} + \mathcal{T}_{\rm comm}^N}.
\end{equation}
On the other hand, Pipe-SGD decouples the dependence of computation and communication by using a stale gradient, but the communication complexity is still $O(T)$ as it needs to communicate all gradients.
Accordingly, its time speedup is
\begin{equation}\label{TS_Pipe-SGD}
{\rm TS}_{\rm Pipe\textrm{-}SGD}^N = \frac{N \cdot \mathcal{T}_{\rm comp}}{\mathcal{T}_{\rm comp} + \mathcal{T}_{\rm comm}^N \cdot a\%}.
\end{equation}
According to Corollary~1 in \citep{yu2018parallel} and Remark~\ref{remark_3}, Local-SGD and CoCoD-SGD have the same convergence rate as S-SGD when the communication period is $O(T^{\frac{1}{4}} / N^{\frac{3}{4}})$.
The difference between Local-SGD and CoCoD-SGD is that CoCoD-SGD runs computation and communication in parallel while Local-SGD runs them sequentially. As a result, their time speedups are
\begin{eqnarray}\label{TS_Local-SGD}
{\rm TS}_{\rm Local\textrm{-}SGD}^N = \frac{N \cdot \mathcal{T}_{\rm comp}}{\mathcal{T}_{\rm comp} + \mathcal{T}_{\rm comm}^N  / k },
\end{eqnarray}
and 
\begin{eqnarray}\label{TS_CoCoD-SGD}
{\rm TS}_{\rm CoCoD\textrm{-}SGD}^N  = \frac{N \cdot \mathcal{T}_{\rm comp}}{\mathcal{T}_{\rm comp} + (\mathcal{T}_{\rm comm}^N \cdot a\%) / k },
\end{eqnarray}
respectively, where $k = O (T^{\frac{1}{4}} / N^{\frac{3}{4}})$. As (\ref{TS_CoCoD-SGD}) is bigger than (\ref{TS_S-SGD}), (\ref{TS_Pipe-SGD}), and (\ref{TS_Local-SGD}), we can verify that CoCoD-SGD achieves the best time speedup.

\section{Experiments}
In this section, we validate the performance of CoCoD-SGD in both homogeneous and heterogeneous environments.

\subsection{Experimental Settings}
\paragraph{Hardware.} 
We evaluate CoCoD-SGD on a cluster where each node has 3 Nvidia Geforce GTX 1080Ti GPUs, 2 Xeon(R) E5-2620 cores and 64 GB memory. The cluster has 6 nodes, which are connected with a 56Gbps InfiniBand network.
Each GPU is viewed as one worker in our experiments.

\paragraph{Software.}
We use Pytorch~0.4.1~\citep{paszke2017automatic} to implement the algorithms in our experiments, and use Horovod~0.15.2~\citep{sergeev2018horovod}, OpenMPI~3.1.2\footnote[1]{https://openmp.org}, and NCCL~2.3.7\footnote[4]{N Luehr. Fast multi-gpu collectives with nccl, 2016.}
to conduct the GPUDirect communication with the Ring-AllReduce algorithm.

\paragraph{Datasets.}
We use two datasets for image classification.
\begin{itemize}
  \item CIFAR10~\citep{krizhevsky2009learning}: it consists of a training set of 50, 000 images from 10 classes, and a test set of 10, 000 images.
  \item CIFAR100~\citep{krizhevsky2009learning}: it is similar to CIFAR10 but has 100 classes.
\end{itemize}

\paragraph{Tasks.} We train ResNet18~\citep{he2016deep} and VGG16~\citep{simonyan2014very} on the two datasets.

\paragraph{Baselines.} We compare CoCoD-SGD with S-SGD, Pipe-SGD~\citep{li2018pipe} and Local-SGD~\citep{stich2018local}.
All of them support Ring-AllReduce communication.

\begin{table}[!b]
\centering
\setlength{\tabcolsep}{1.5mm}{
\begin{tabular}{|l|l|l|l|l|}
\hline
& \multicolumn{2}{|c|}{CIFAR10} & \multicolumn{2}{|c|}{CIFAR100} \\
\hline
& ResNet18 & VGG16 & ResNet18 & VGG16 \\ 
\hline
S-SGD & 94.00\% & 93.25\% & 75.08\% & 70.81\% \\
\hline
Pipe-SGD & 93.94\% & 93.09\% & 75.13\% & 70.59\%    \\
\hline
Local-SGD & 94.35\% & 93.30\% & 75.42\% & 71.13\%    \\
\hline
CoCoD-SGD & 94.41\% & 93.38\% & 75.67\% & 72.24\%    \\
\hline
\end{tabular}}
\caption{Final best test accuracy for all tasks in a homogeneous environment. 16 workers in total.}
\label{final_accuracy}
\end{table}

\begin{figure*}[t]
\centering

\subfigure[{\tiny CIFAR10-ResNet-Epoch-Loss}]{
\begin{minipage}[t]{0.225\linewidth}
\centering
\includegraphics[width=1.1\textwidth] {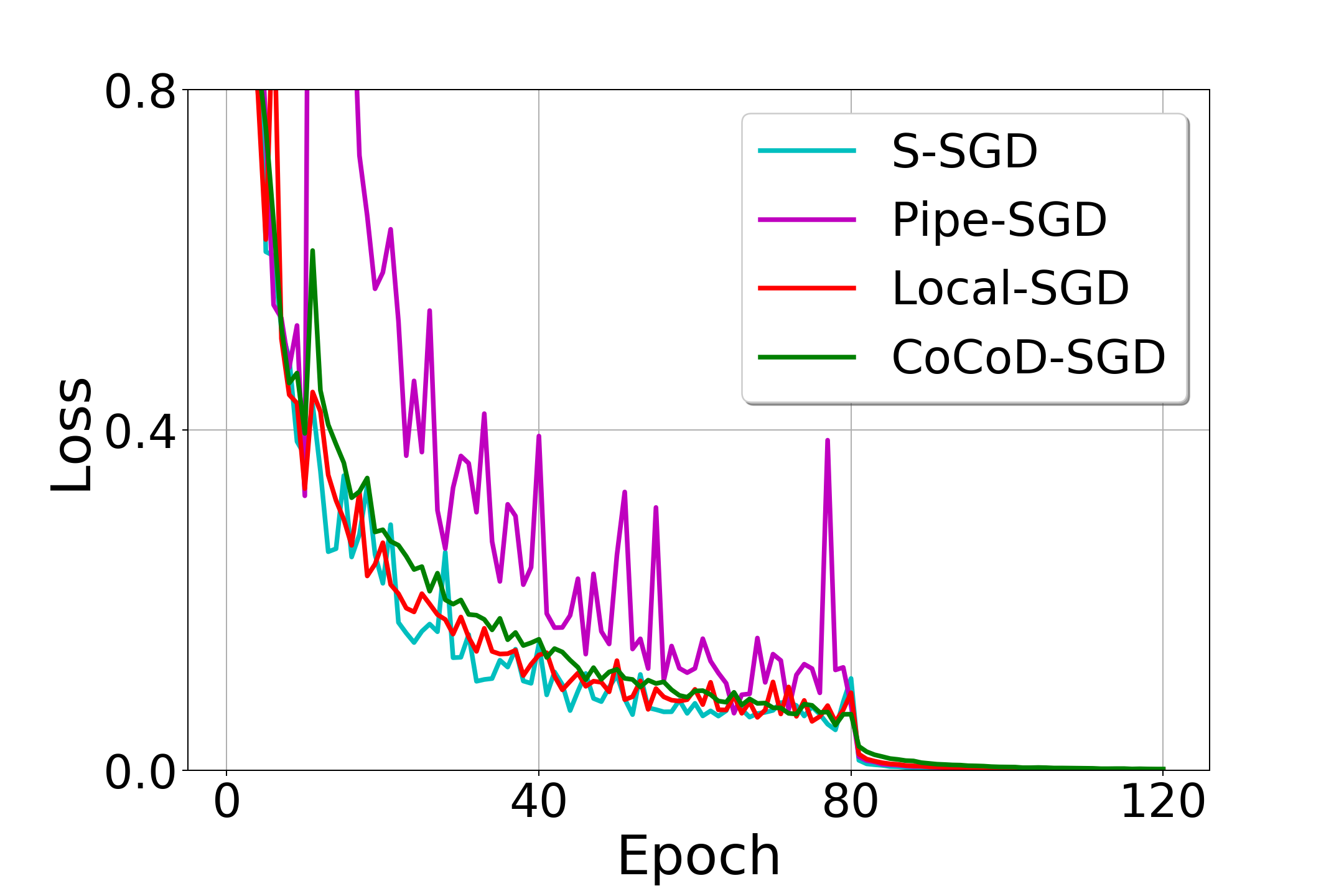}
\end{minipage}
}
\subfigure[{\tiny CIFAR10-VGG-Epoch-Loss}]{
\begin{minipage}[t]{0.225\linewidth}
\centering
\includegraphics[width=1.1\textwidth] {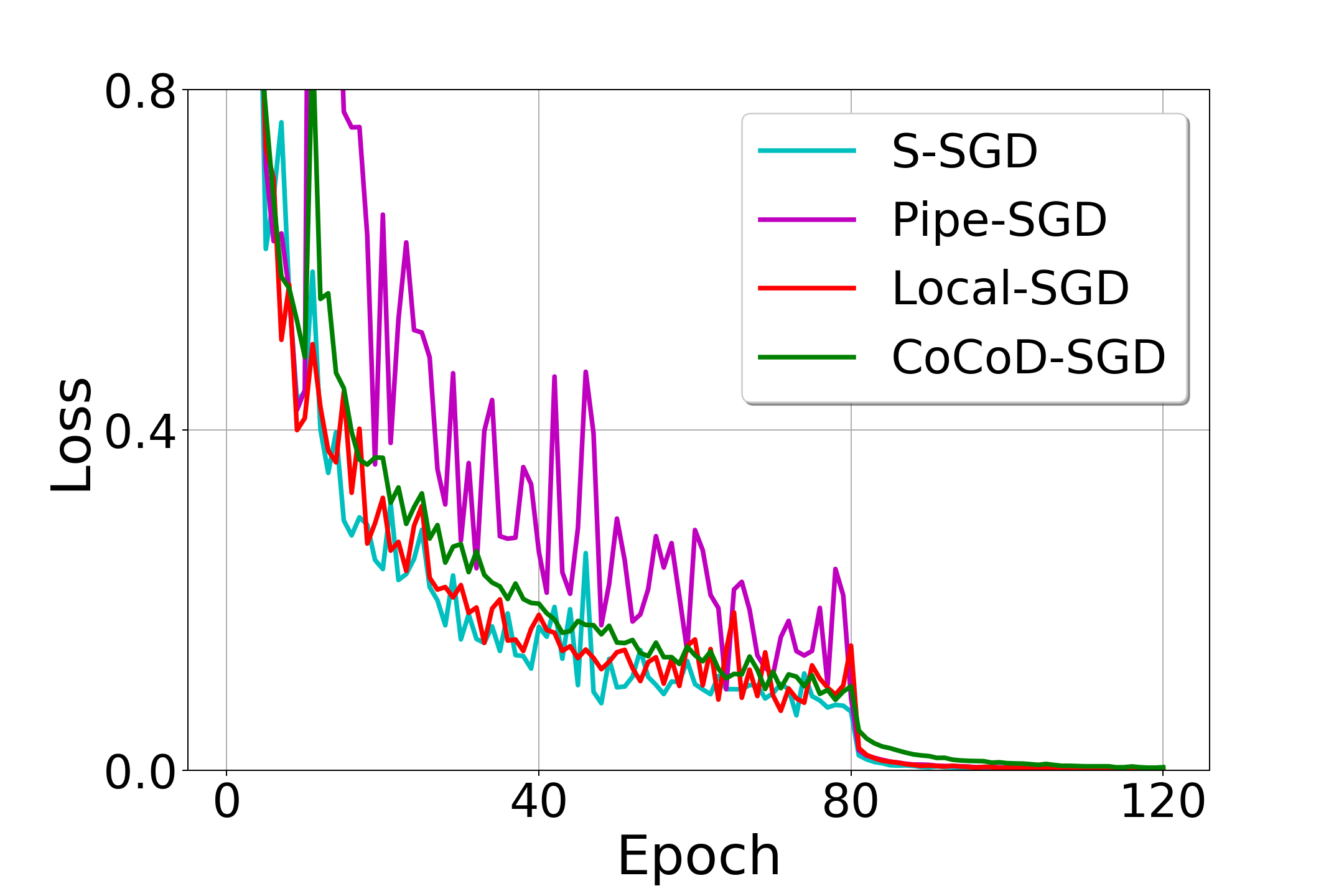}
\end{minipage}
}
\subfigure[{\tiny CIFAR100-ResNet-Epoch-Loss}]{
\begin{minipage}[t]{0.225\linewidth}
\centering
\includegraphics[width=1.1\textwidth] {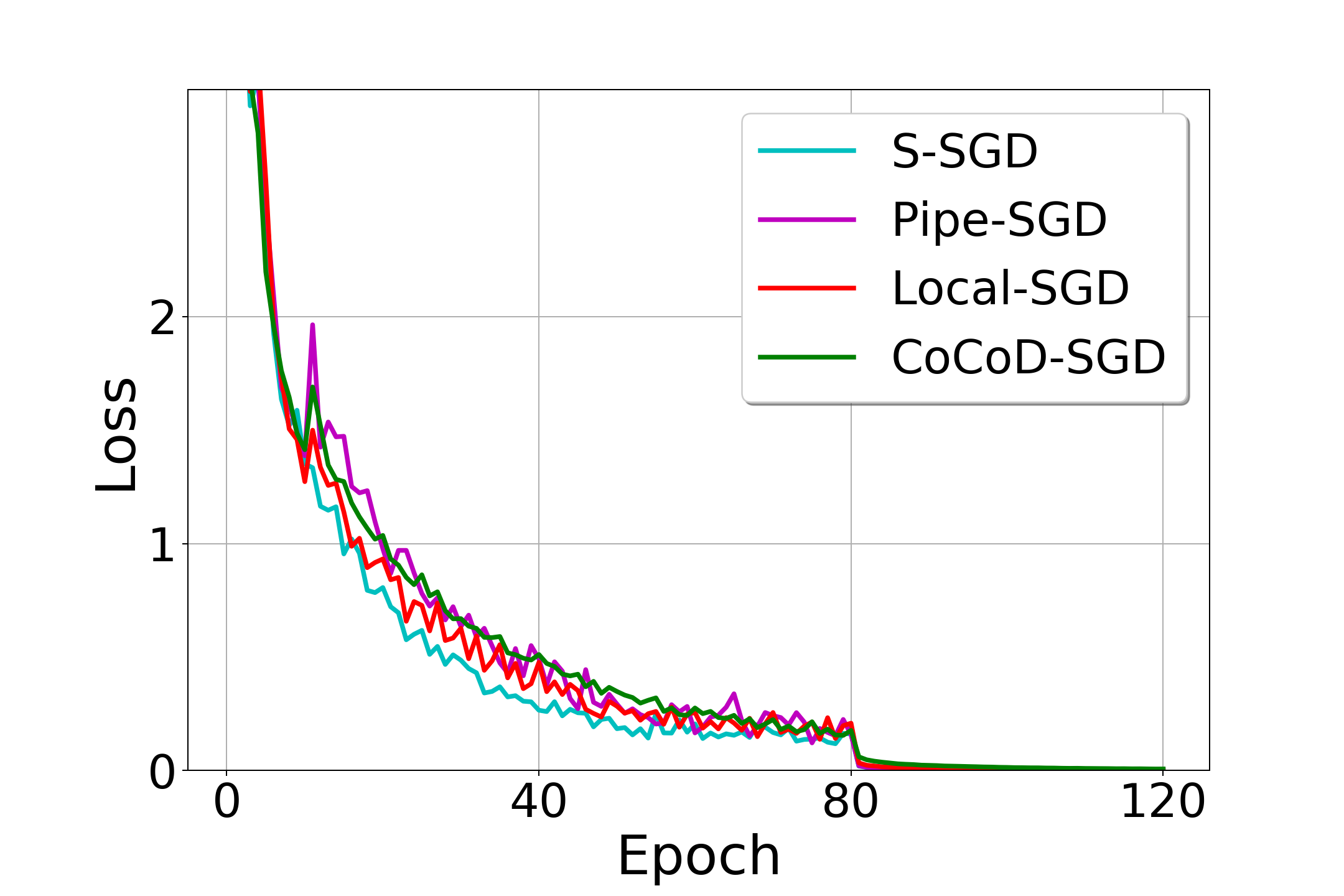}
\end{minipage}
}
\subfigure[{\tiny CIFAR100-VGG-Epoch-Loss}]{
\begin{minipage}[t]{0.225\linewidth}
\centering
\includegraphics[width=1.1\textwidth] {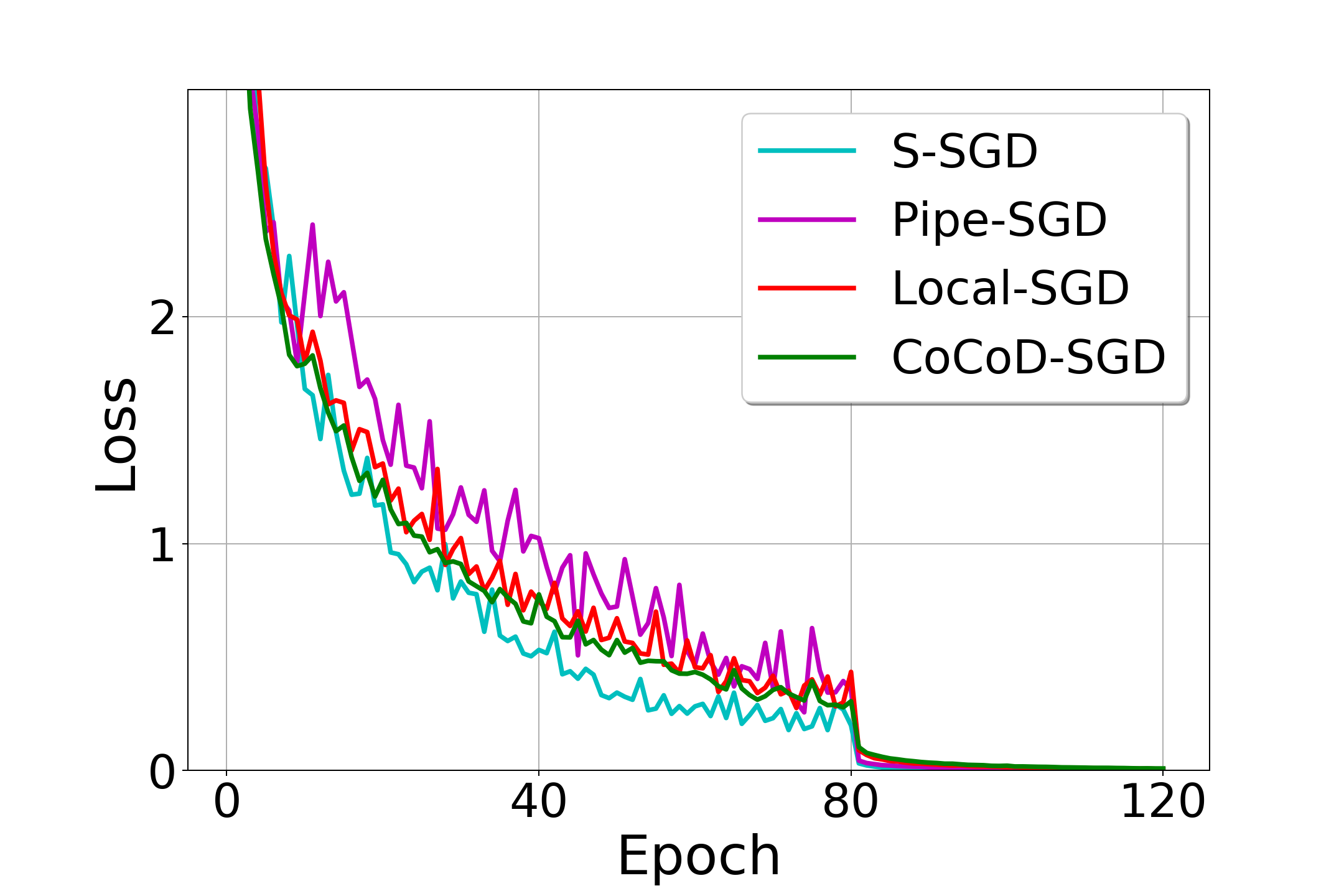}
\end{minipage}
}
\caption{Training loss for ResNet18 and VGG16 on CIFAR10 and CIFAR100 w.r.t epochs in a homogeneous environment. All algorithms have a similar convergence rate.}
\label{Compare_epoch}
\end{figure*}

\begin{figure*}[t]
\centering
\subfigure[{\tiny CIFAR10-ResNet-Time-Accuracy}]{
\begin{minipage}[t]{0.225\linewidth}
\centering
\includegraphics[width=1.1\textwidth] {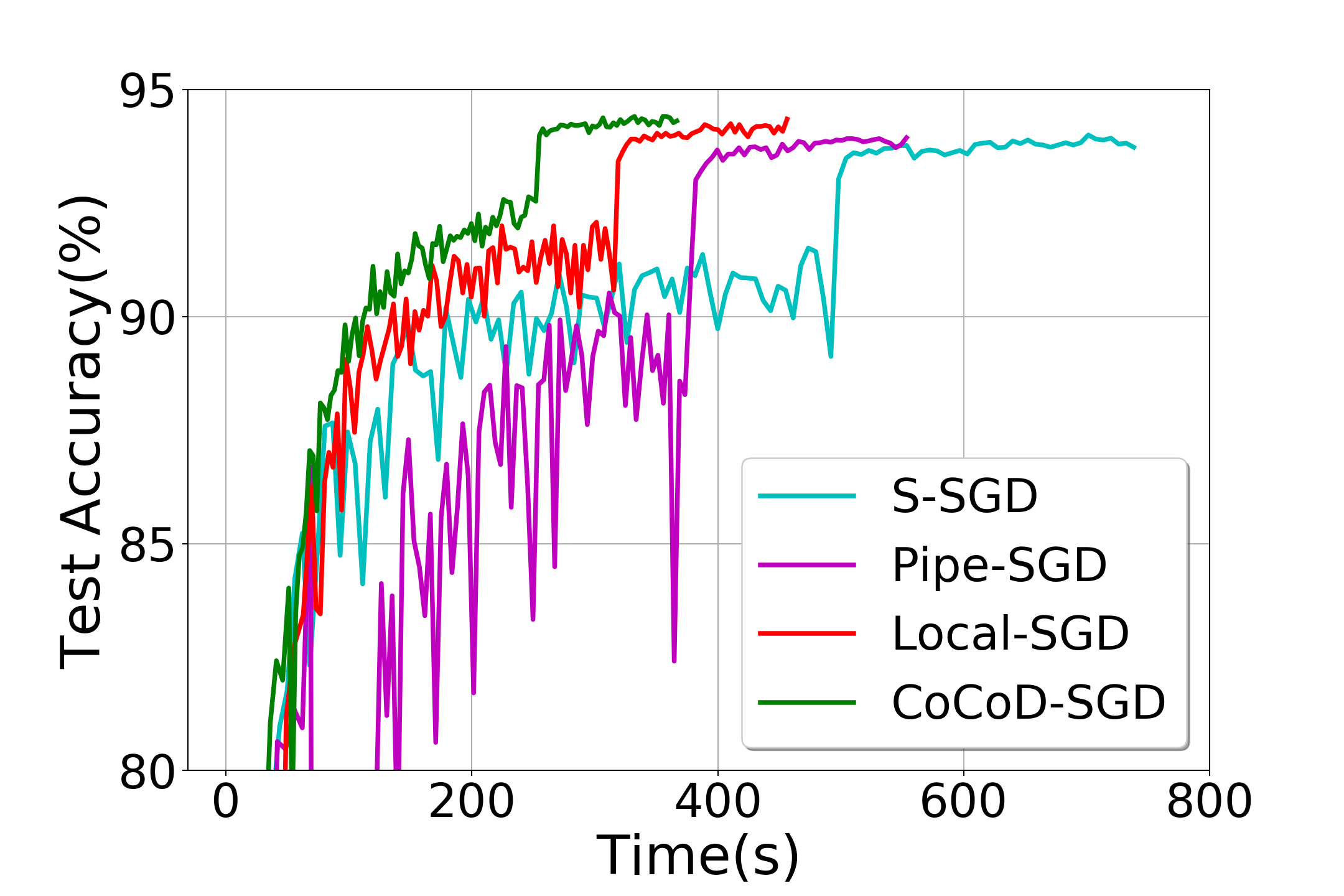}
\end{minipage}
}
\subfigure[{\tiny CIFAR10-VGG-Time-Accuracy}]{
\begin{minipage}[t]{0.225\linewidth}
\centering
\includegraphics[width=1.1\textwidth] {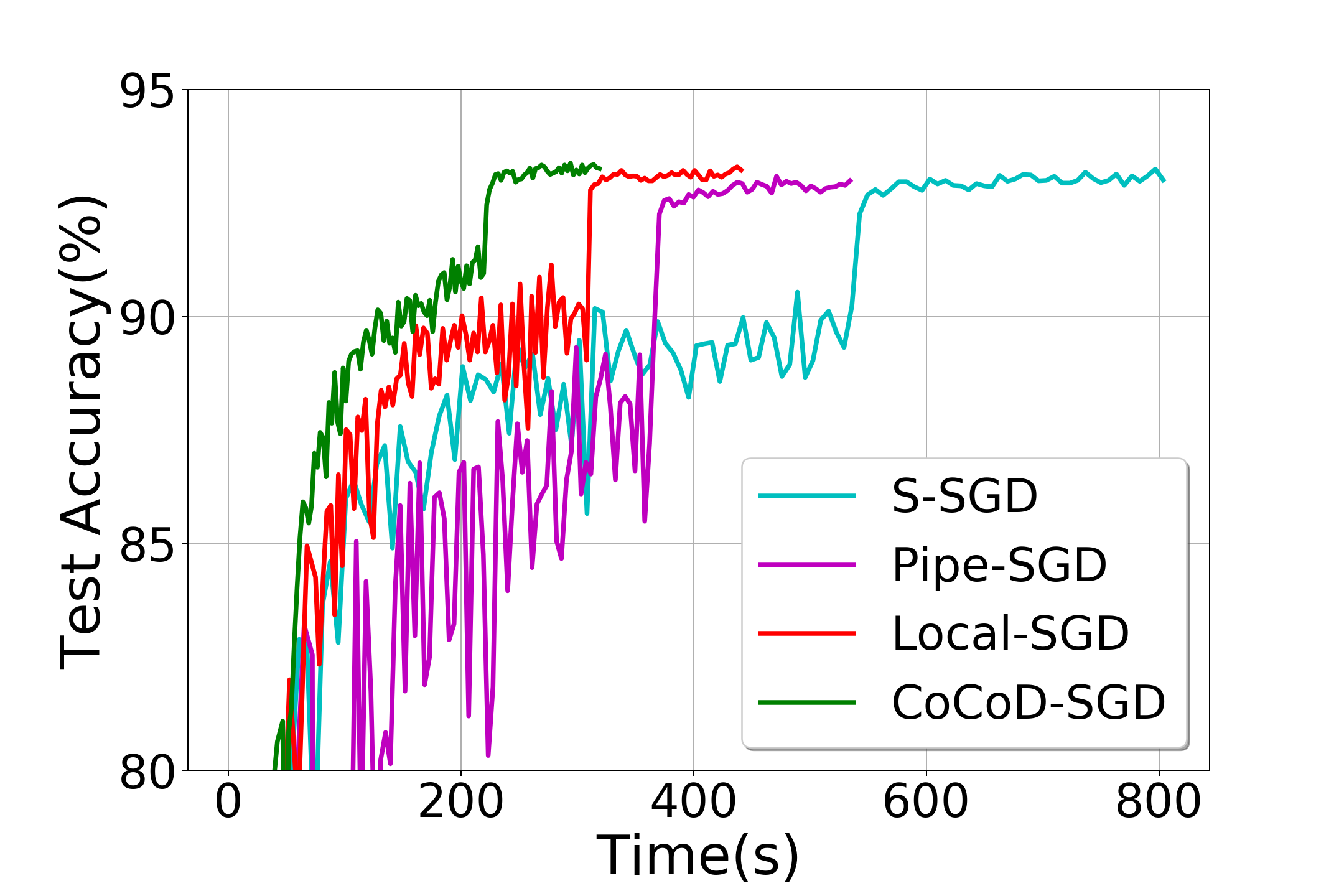}
\end{minipage}
}
\subfigure[{\tiny CIFAR100-ResNet-Time-Accuracy}]{
\begin{minipage}[t]{0.225\linewidth}
\centering
\includegraphics[width=1.1\textwidth] {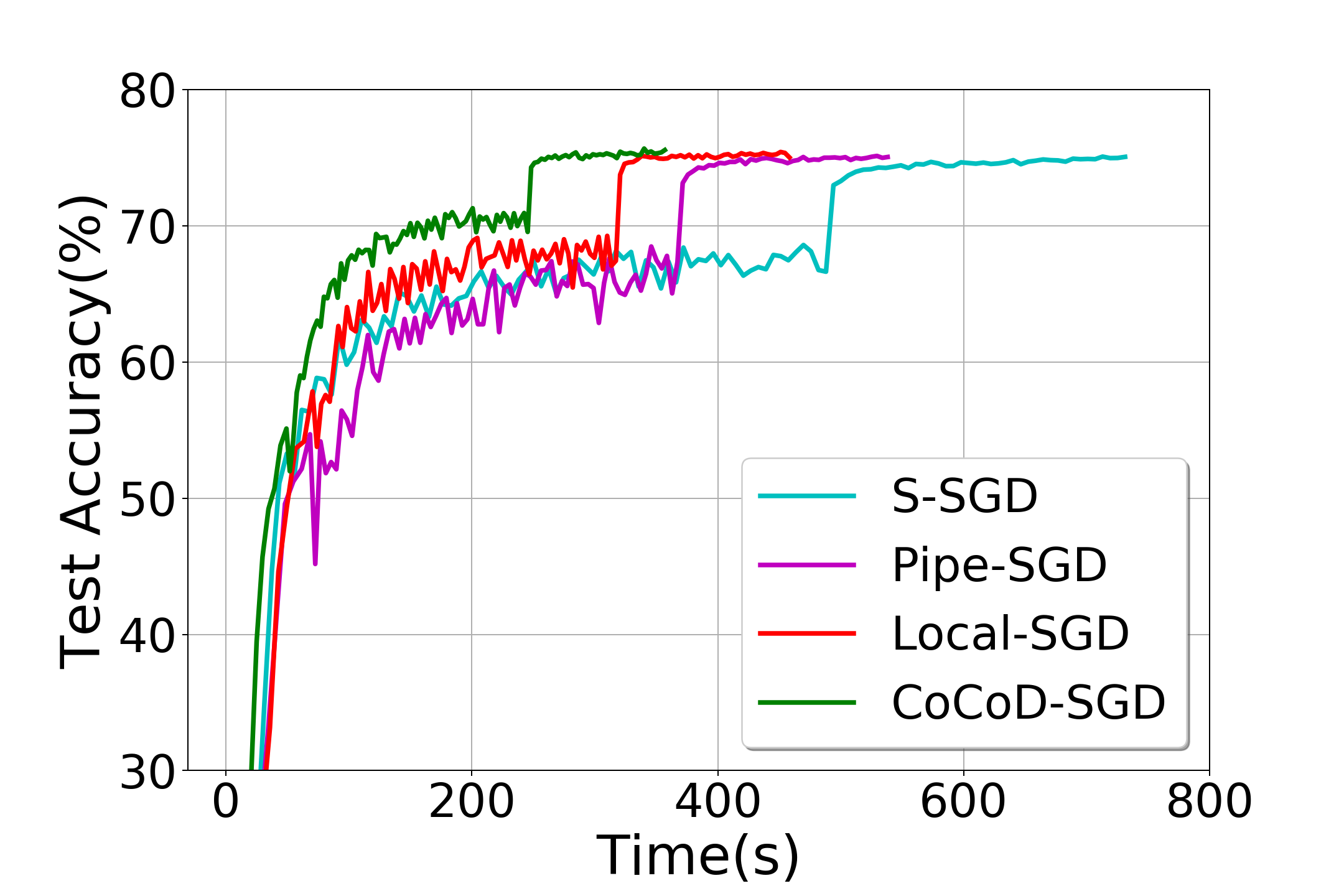}
\end{minipage}
}
\subfigure[{\tiny CIFAR100-VGG-Time-Accuracy}]{
\begin{minipage}[t]{0.225\linewidth}
\centering
\includegraphics[width=1.1\textwidth] {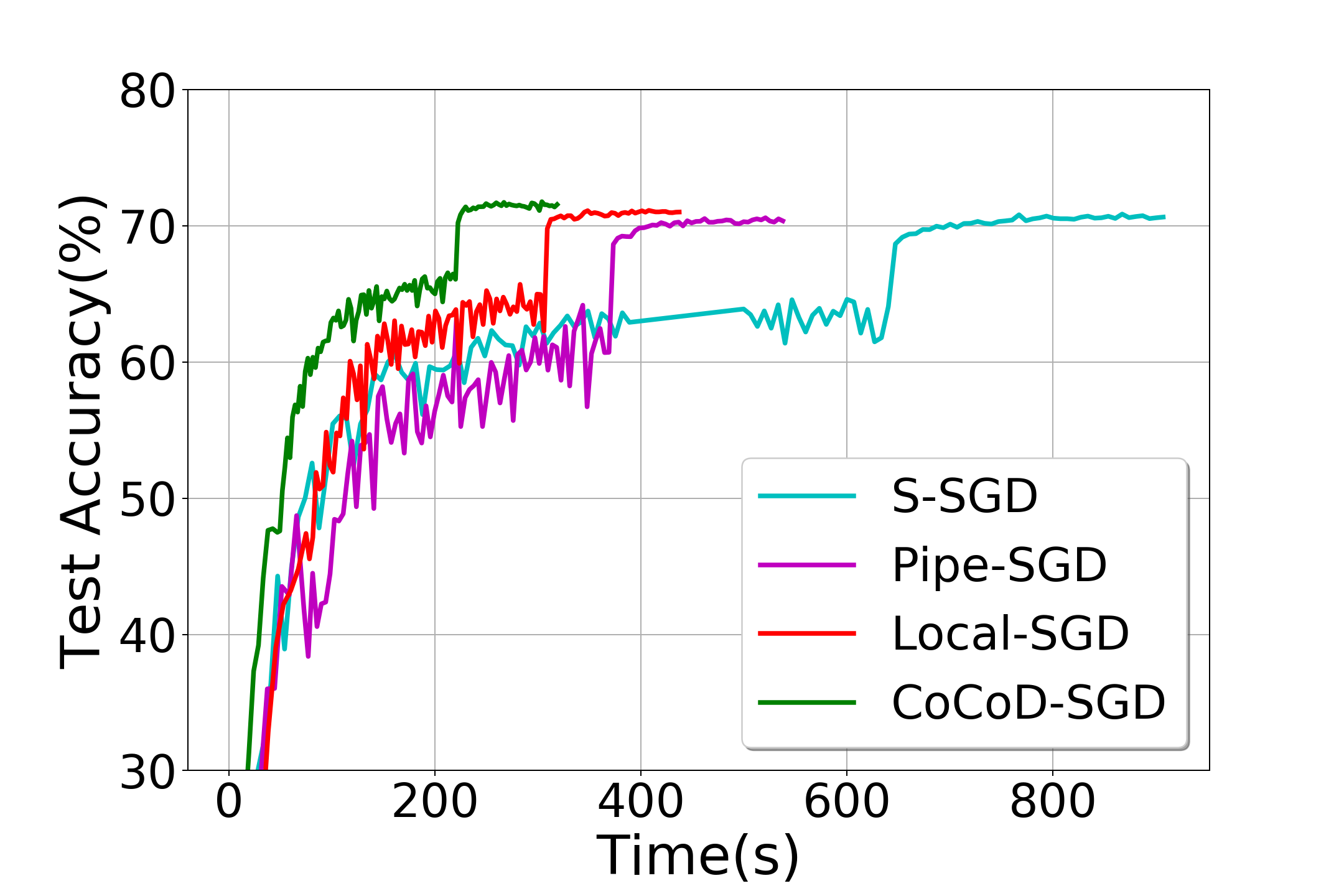}
\end{minipage}
}
\caption{Test accuracy for ResNet18 and VGG16 on CIFAR10 and CIFAR100 w.r.t time in a homogeneous environment. CoCoD-SGD achieves the fastest convergence and the results are consistent with the analysis in Section 4.2.}
\label{Compare_time}
\end{figure*}

\begin{figure*}[!t]
\centering
\subfigure[{\tiny CIFAR10-ResNet-Speedup}]{
\begin{minipage}[t]{0.225\linewidth}
\centering
\includegraphics[width=1.1\textwidth] {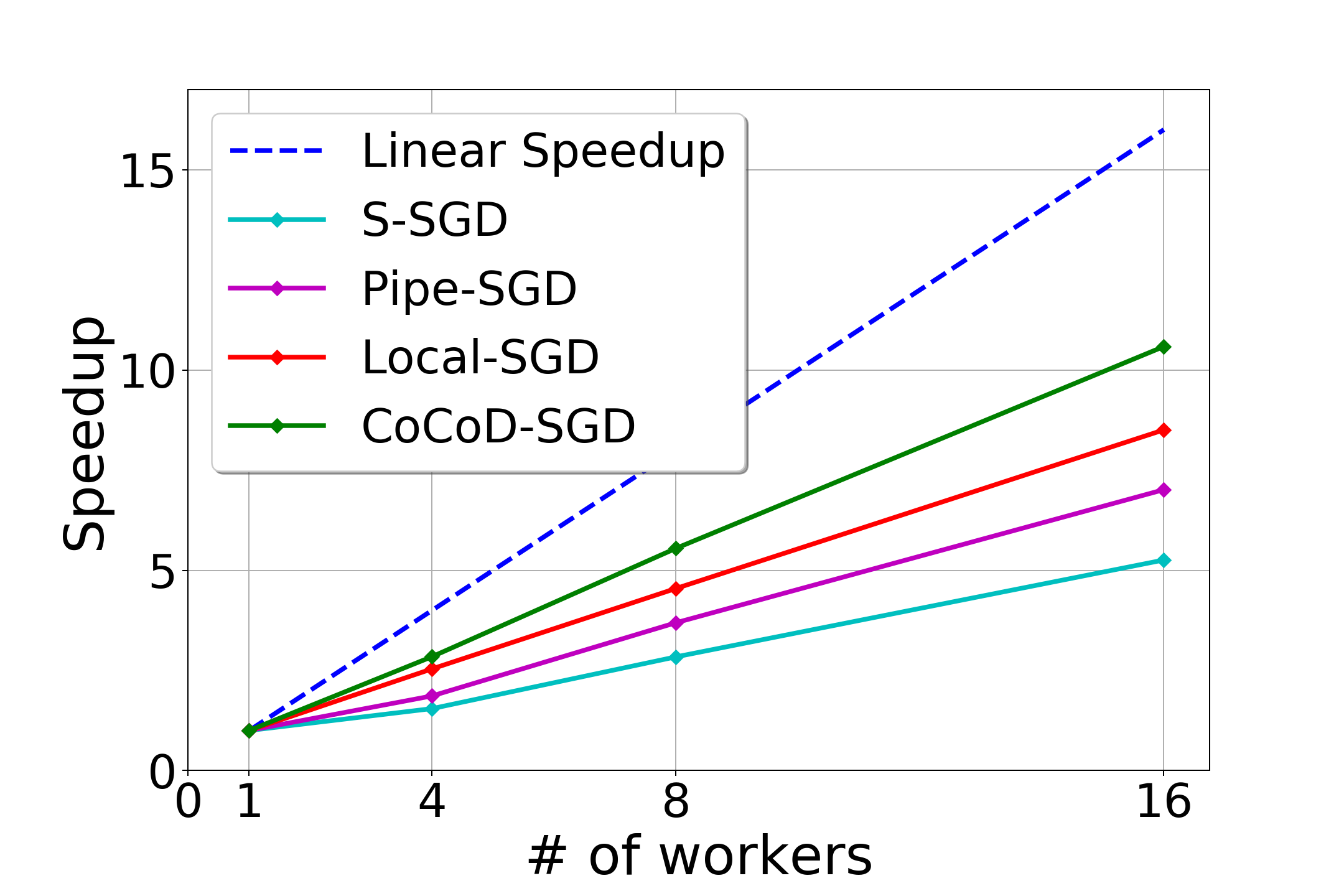}
\end{minipage}
}
\subfigure[{\tiny CIFAR10-VGG-Speedup}]{
\begin{minipage}[t]{0.225\linewidth}
\centering
\includegraphics[width=1.1\textwidth] {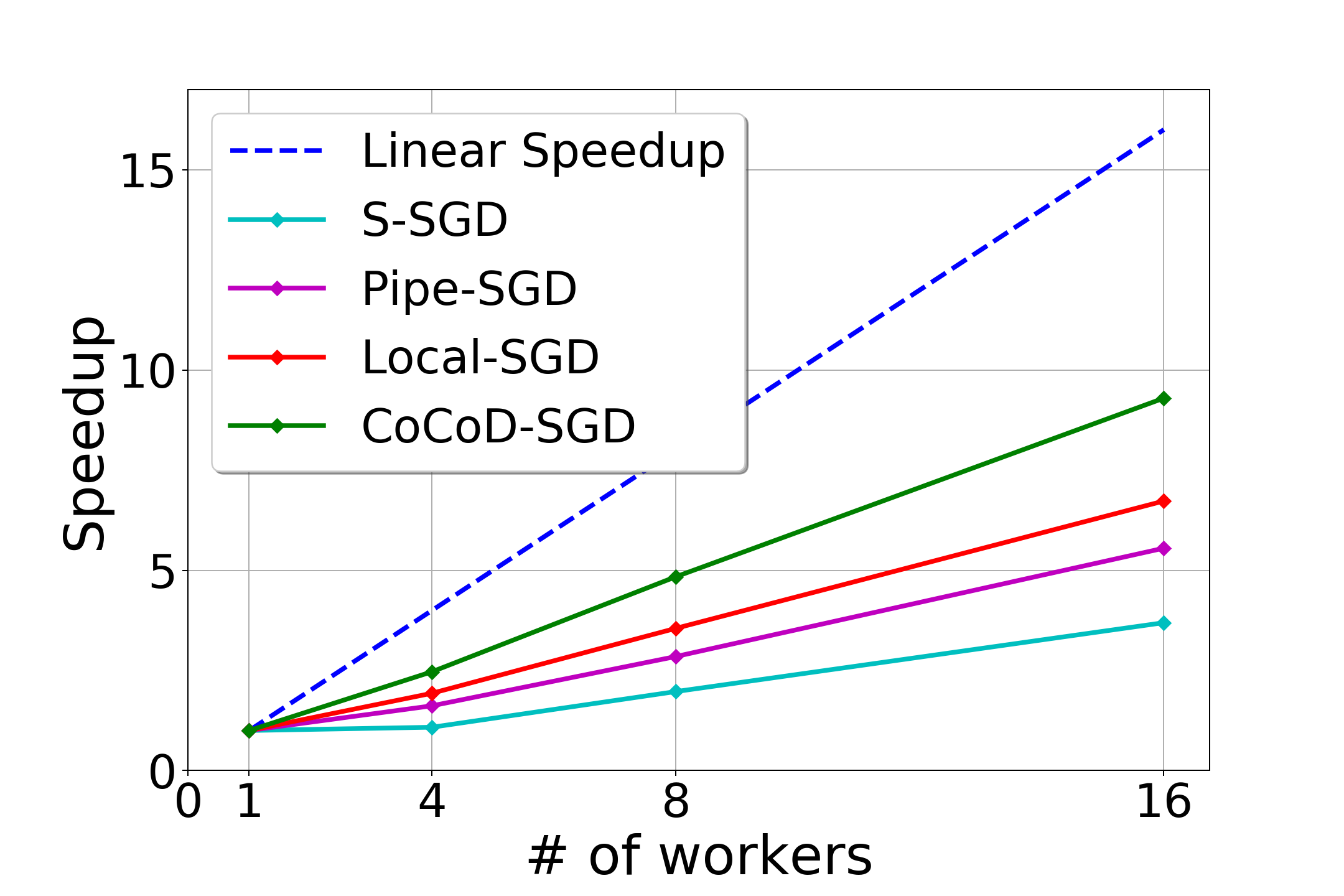}
\end{minipage}
}
\subfigure[{\tiny CIFAR100-ResNet-Speedup}]{
\begin{minipage}[t]{0.225\linewidth}
\centering
\includegraphics[width=1.1\textwidth] {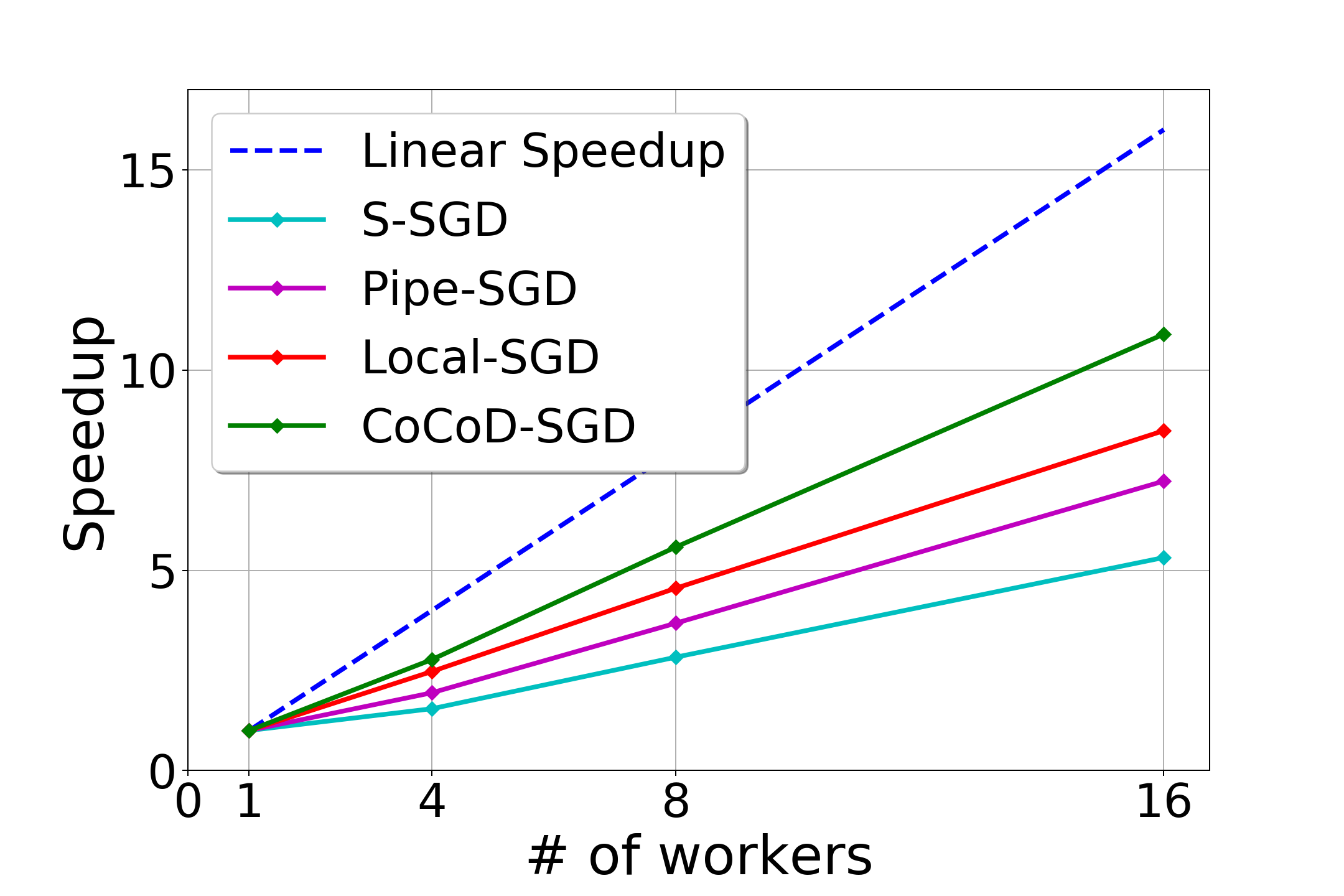}
\end{minipage}
}
\subfigure[{\tiny CIFAR100-VGG-Speedup}]{
\begin{minipage}[t]{0.225\linewidth}
\centering
\includegraphics[width=1.1\textwidth] {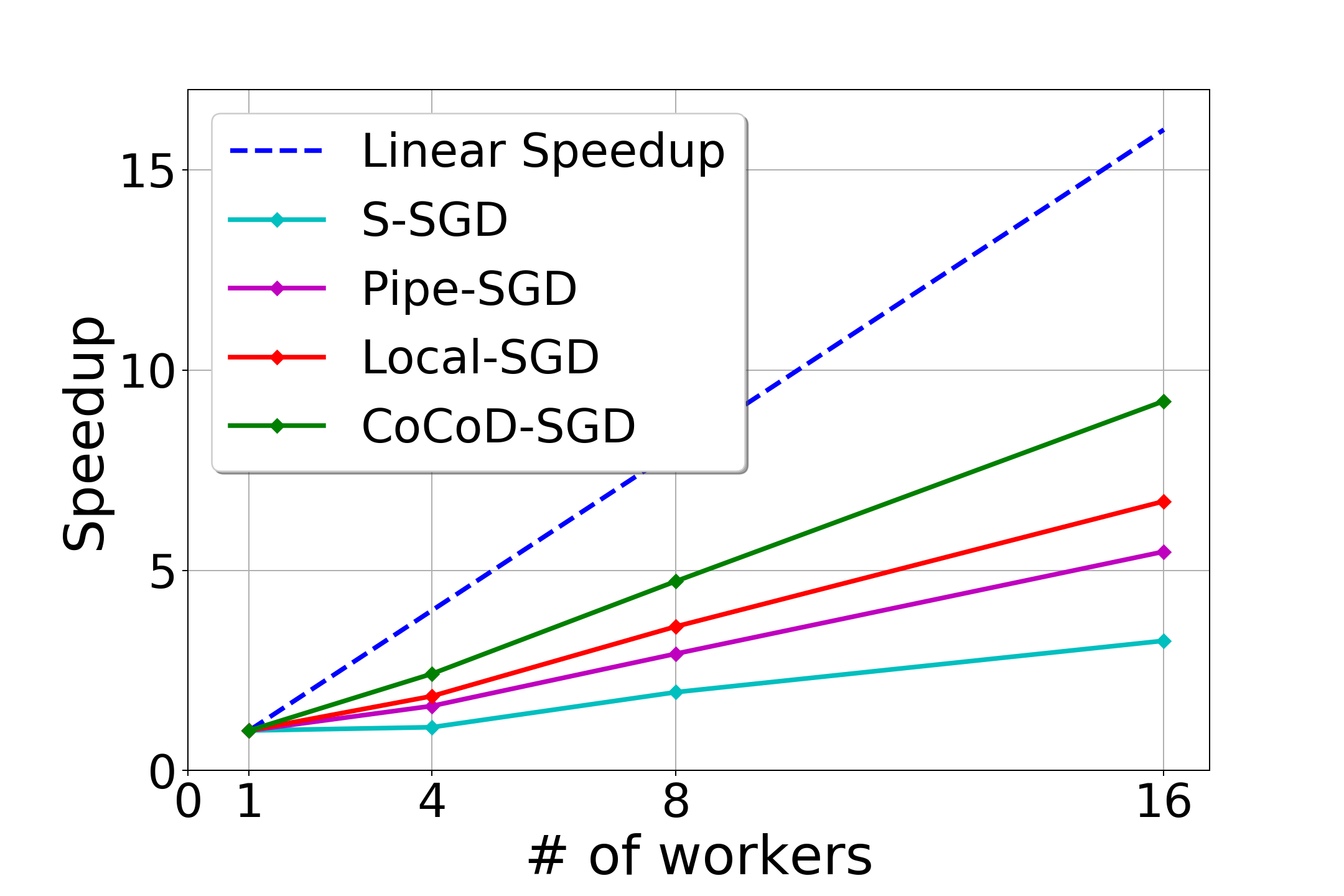}
\end{minipage}
}
\caption{The comparison of time speedup for ResNet18 and VGG16 on CIFAR10 and CIFAR100 in a homogeneous environment. CoCoD-SGD achieves the best time speedup.}
\label{Compare_speedup}
\end{figure*}

\paragraph{Hyper-parameters.} We use the following hyper-parameters.
\begin{itemize}
  \item Basic batch size: 32 for both ResNet18 and VGG16.
  \item Basic learning rate: For both networks we start the learning rate from 0.01 and decay it by a factor of 10 at the beginning of the 81st epoch. 
  \item Momentum: 0.9.
  \item Weight decay: $10^{-4}$.
  \item Communication period and gradient staleness: Since the variance of stochastic gradients is higher in the beginning, we set the communication period to 1 for the first 10 epochs and 5 for 
  the subsequential epochs. 
  The staleness of gradients in Pipe-SGD is set to 1 as suggested in \citep{li2018pipe}.
\end{itemize}

\subsection{Homogeneous Environment}
In a homogeneous environment, all workers have the same computation speed. 
So for $N$ workers, we set $\gamma_N = N \gamma$ as suggested in Remark~\ref{remark_4}. 
And the learning rate warm-up scheme proposed in~\citep{goyal2017accurate} is adopted.
\paragraph{Comparison of convergence rate.} 
Figure~\ref{Compare_epoch} shows the training loss with regard to epochs of ResNet18 and VGG16 on 16 GPUs. 
All 
algorithms have similar convergence speed, which validates the theoretical results claimed in Section 4.1.
\paragraph{Comparison of convergence speed.} 
Figure~\ref{Compare_time} shows the test accuracy regarding time on 16 GPUs and Table~\ref{final_accuracy} shows the best test accuracies of all algorithms on the two datasets.
We evaluate the accuracy on the test set during training, but we only accumulate the time used for training.
As shown in Figure~\ref{Compare_time}, CoCoD-SGD achieves almost $2\times$ and $2.5\times$ speedup against S-SGD for ResNet18 and VGG16 respectively. 
Although Pipe-SGD can run computation and communication in parallel, it is still slower than Local-SGD and CoCoD-SGD since its communication complexity is higher. 
CoCoD-SGD converges faster than others since it not only runs computation and communication simultaneously but also has a lower communication complexity. 
The results verify our theoretical results claimed in Section 4.2. 
In the meanwhile, we can observe from Table~\ref{final_accuracy} that CoCoD-SGD does not sacrifice the test accuracy 
on both datasets 
and may get better results than S-SGD.
\paragraph{Comparison of time speedup.} 
Figure~\ref{Compare_speedup} shows the time speedup for ResNet18 and VGG16 when the number of workers increases from 1 to 16. 
We run the experiments for 120 epochs on 1 GPU and multiple GPUs. 
The speedup for ResNet18 is better due to its smaller model size.
On both tasks, CoCoD-SGD achieves the fastest convergence and the best time speedup, which validates our time speedup analysis.

\begin{figure*}[!b]
\centering
\subfigure[{\tiny Heter-CIFAR10-ResNet-Epoch-Loss}]{
\begin{minipage}[t]{0.225\linewidth}
\centering
\includegraphics[width=1.1\textwidth] {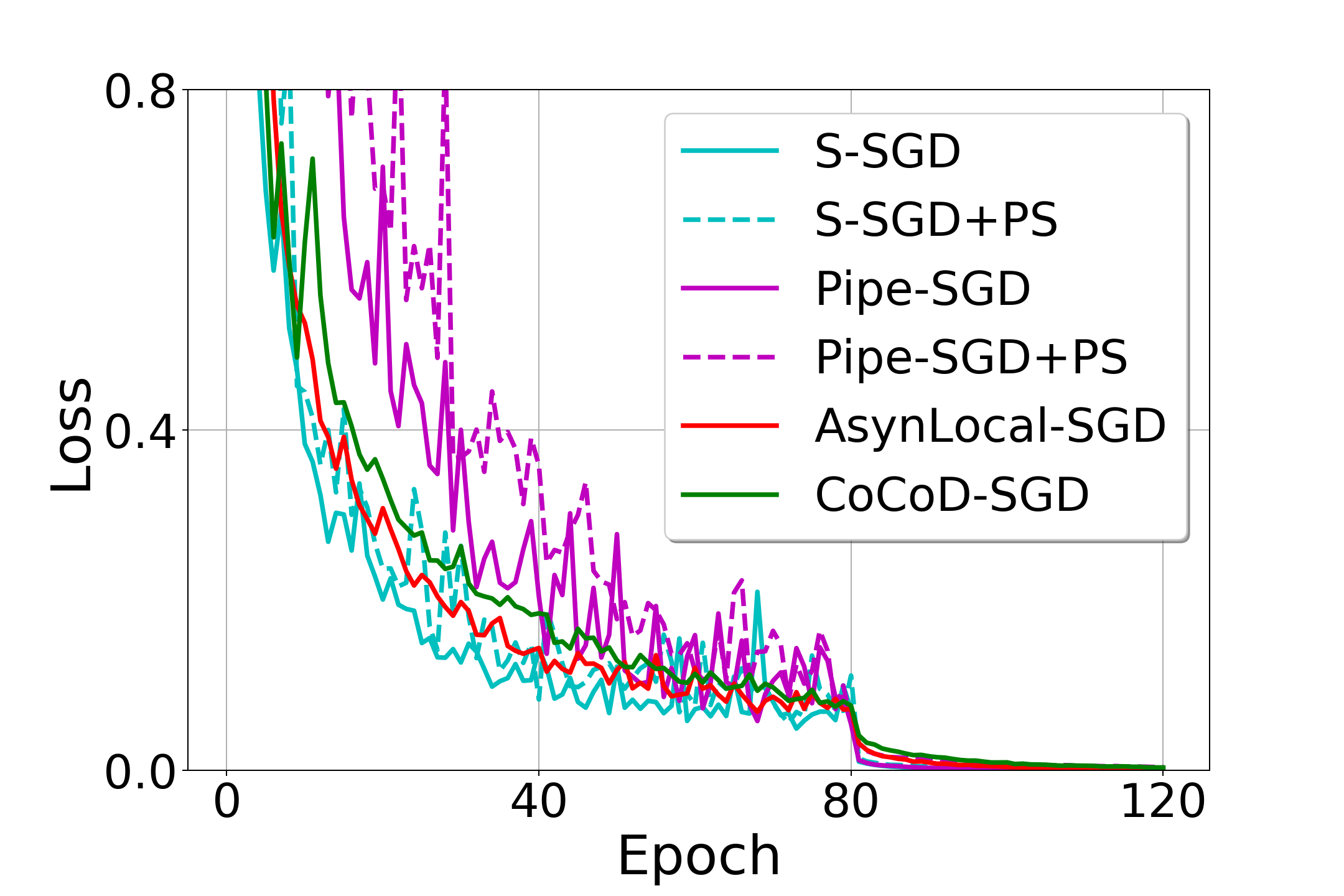}
\end{minipage}
}
\subfigure[{\tiny Heter-CIFAR10-VGG-Epoch-Loss}]{
\begin{minipage}[t]{0.225\linewidth}
\centering
\includegraphics[width=1.1\textwidth] {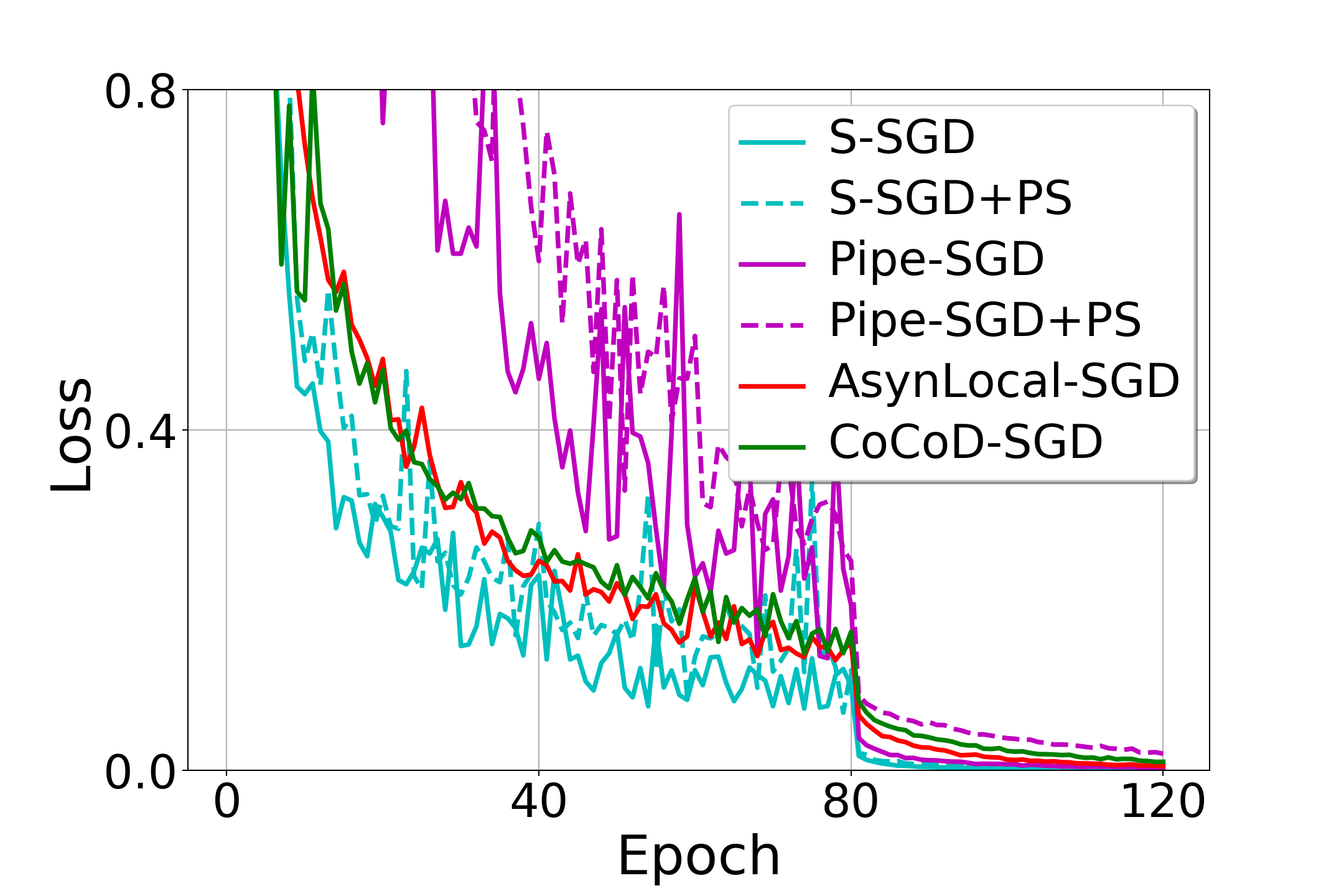}
\end{minipage}
}
\subfigure[{\tiny Heter-CIFAR100-ResNet-Epoch-Loss}]{
\begin{minipage}[t]{0.225\linewidth}
\centering
\includegraphics[width=1.1\textwidth] {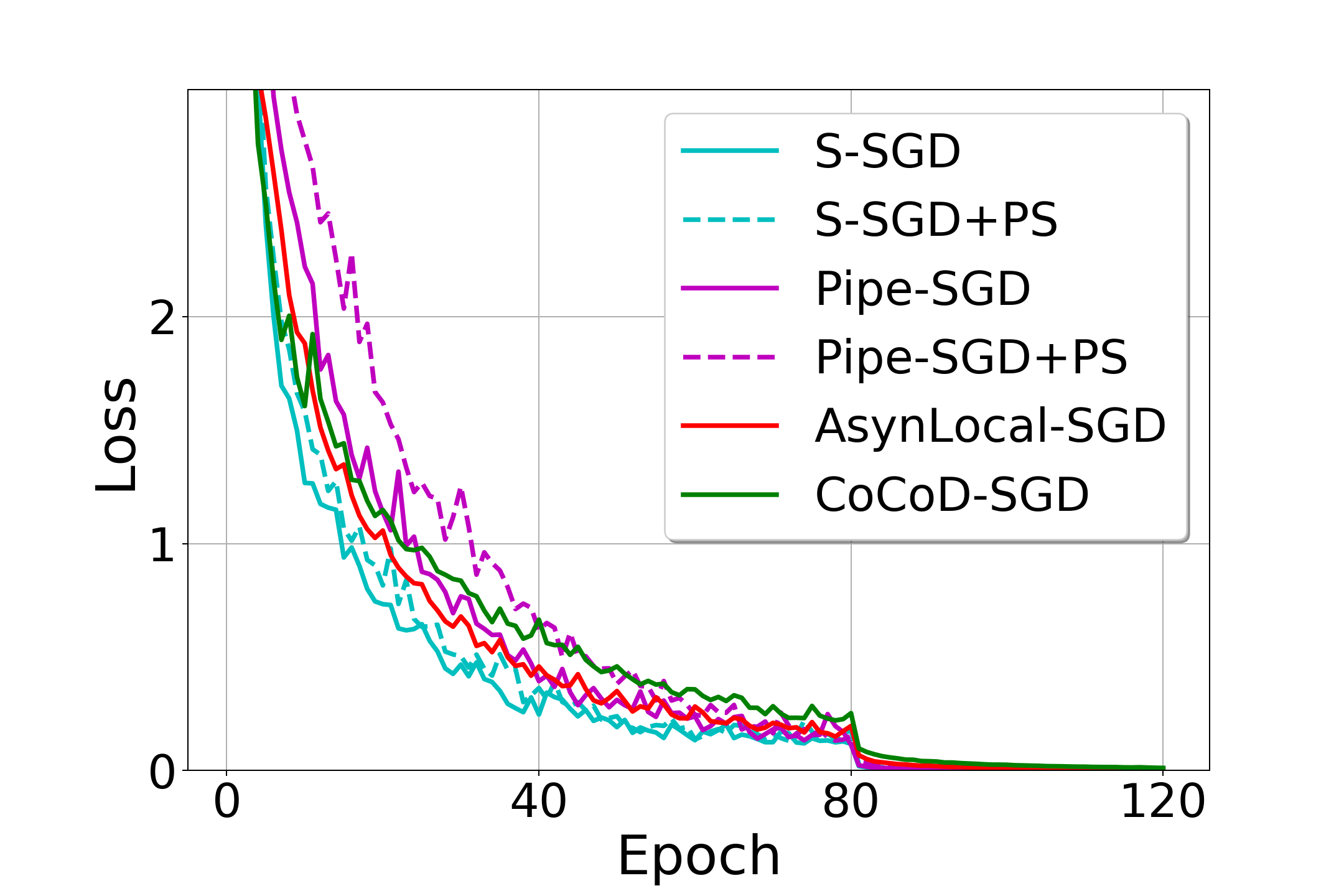}
\end{minipage}
}
\subfigure[{\tiny Heter-CIFAR100-VGG-Epoch-Loss}]{
\begin{minipage}[t]{0.225\linewidth}
\centering
\includegraphics[width=1.1\textwidth] {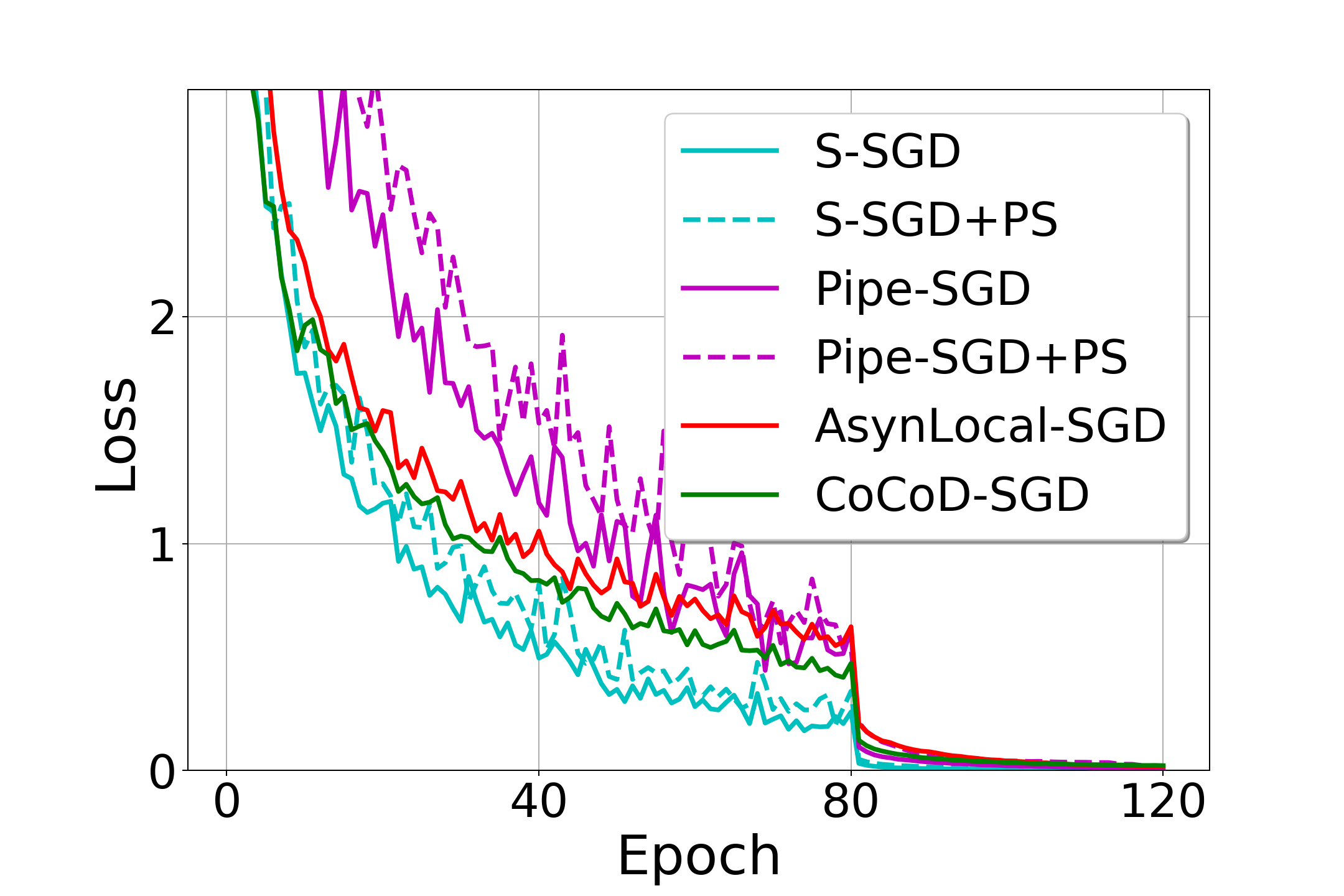}
\end{minipage}
}
\caption{Training loss for ResNet18 and VGG16 on CIFAR10 and CIFAR100 w.r.t epochs in a heterogeneous environment. All algorithms have a similar convergence rate.}
\label{Compare_epoch_heter}
\end{figure*}
 
\begin{figure*}[!b]
\centering
\subfigure[{\tiny Heter-CIFAR10-ResNet-Time-Accuracy}]{
\begin{minipage}[t]{0.225\linewidth}
\centering
\includegraphics[width=1.1\textwidth] {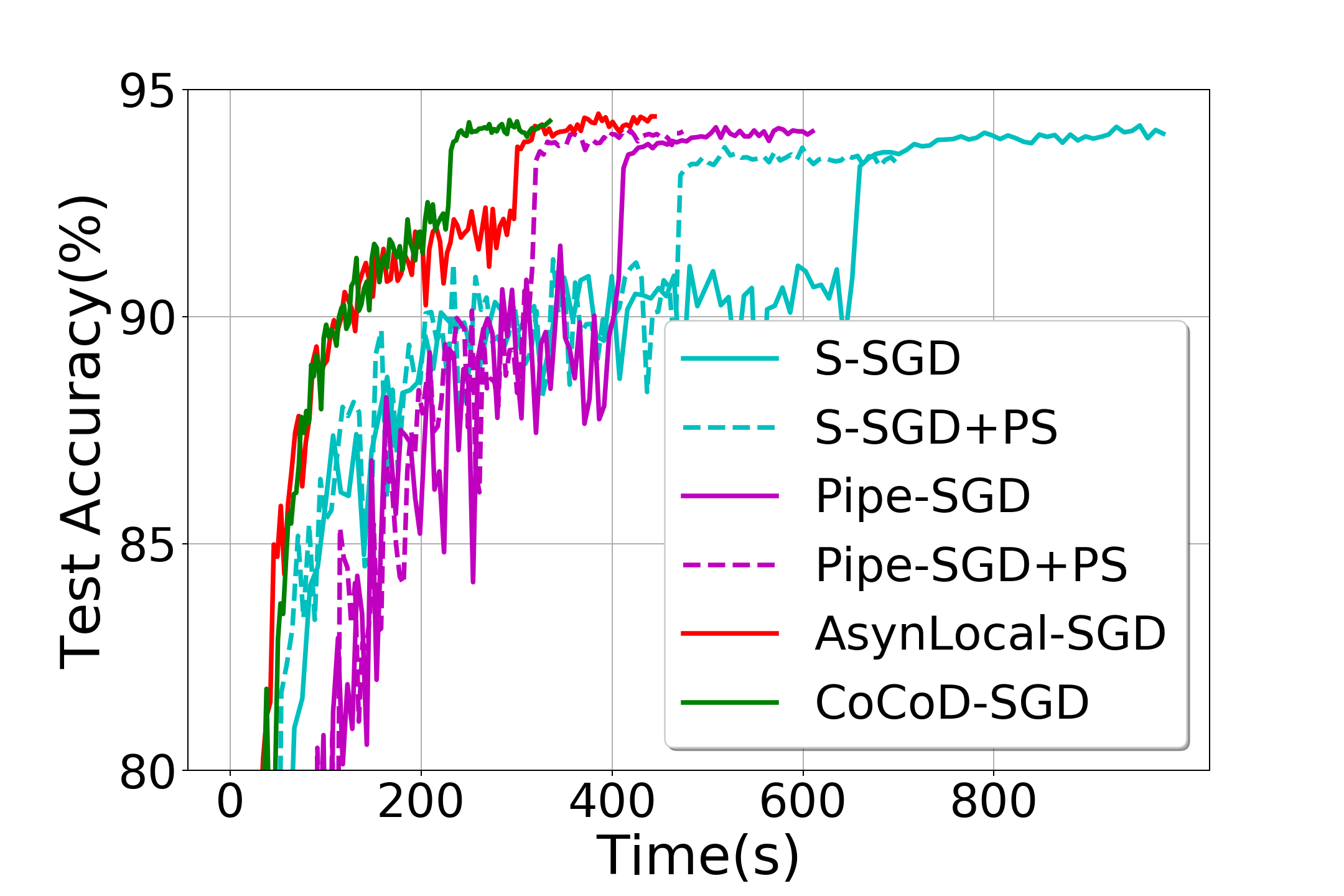}
\end{minipage}
}
\subfigure[{\tiny Heter-CIFAR10-VGG-Time-Accuracy}]{
\begin{minipage}[t]{0.225\linewidth}
\centering
\includegraphics[width=1.1\textwidth] {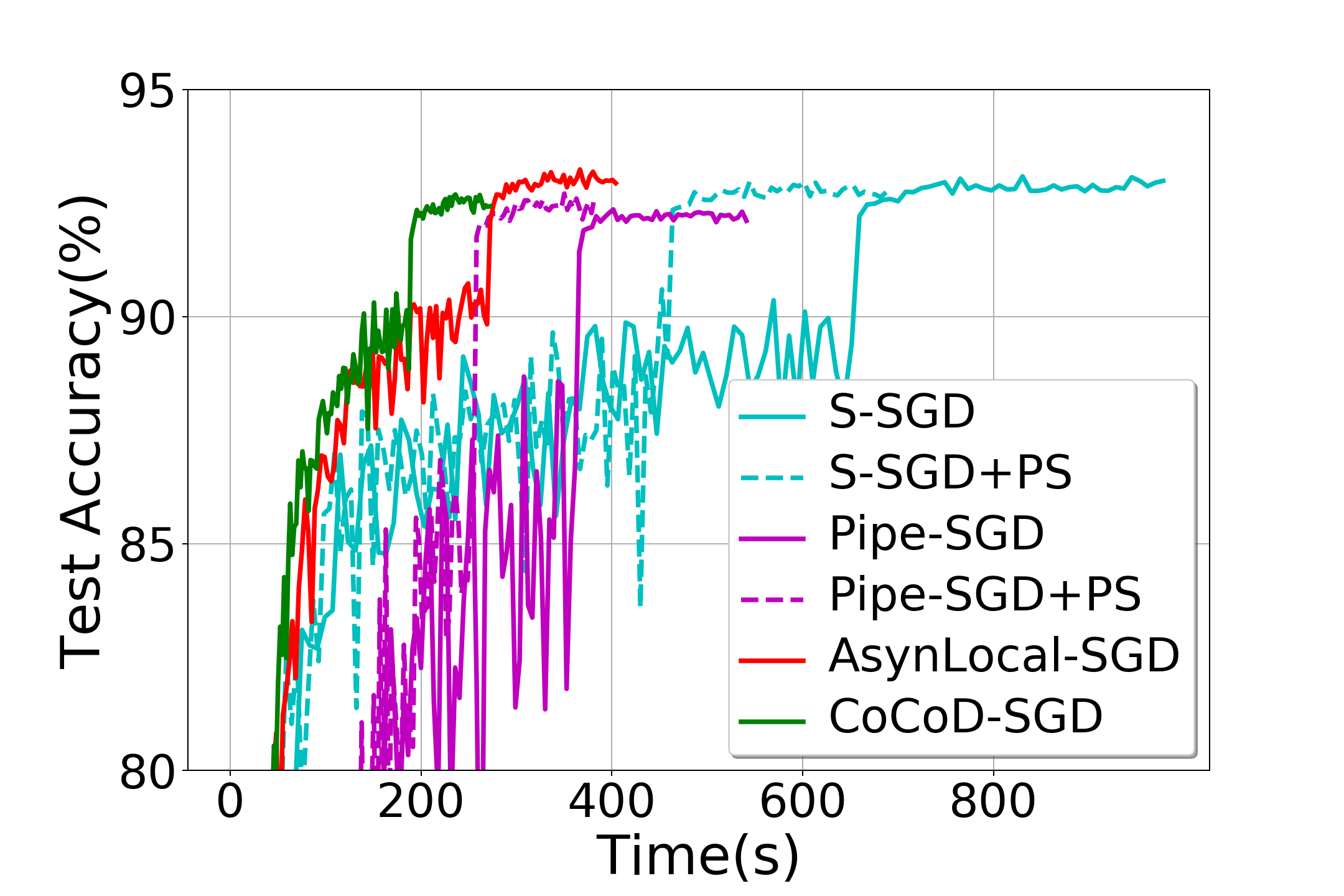}
\end{minipage}
}
\subfigure[{\tiny Heter-CIFAR100-ResNet-Time-Accuracy}]{
\begin{minipage}[t]{0.225\linewidth}
\centering
\includegraphics[width=1.1\textwidth] {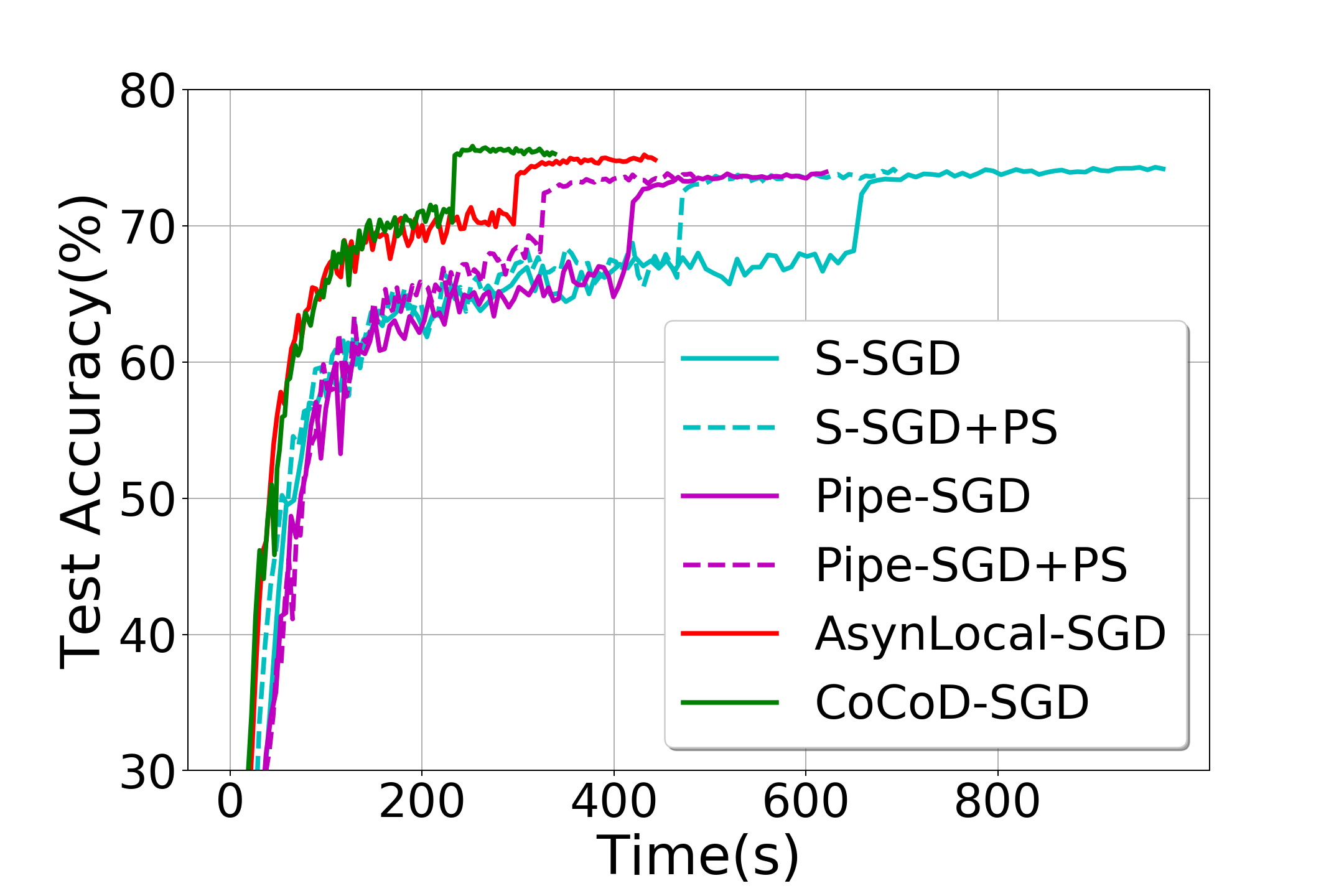}
\end{minipage}
}
\subfigure[{\tiny Heter-CIFAR100-VGG-Time-Accuracy}]{
\begin{minipage}[t]{0.225\linewidth}
\centering
\includegraphics[width=1.1\textwidth] {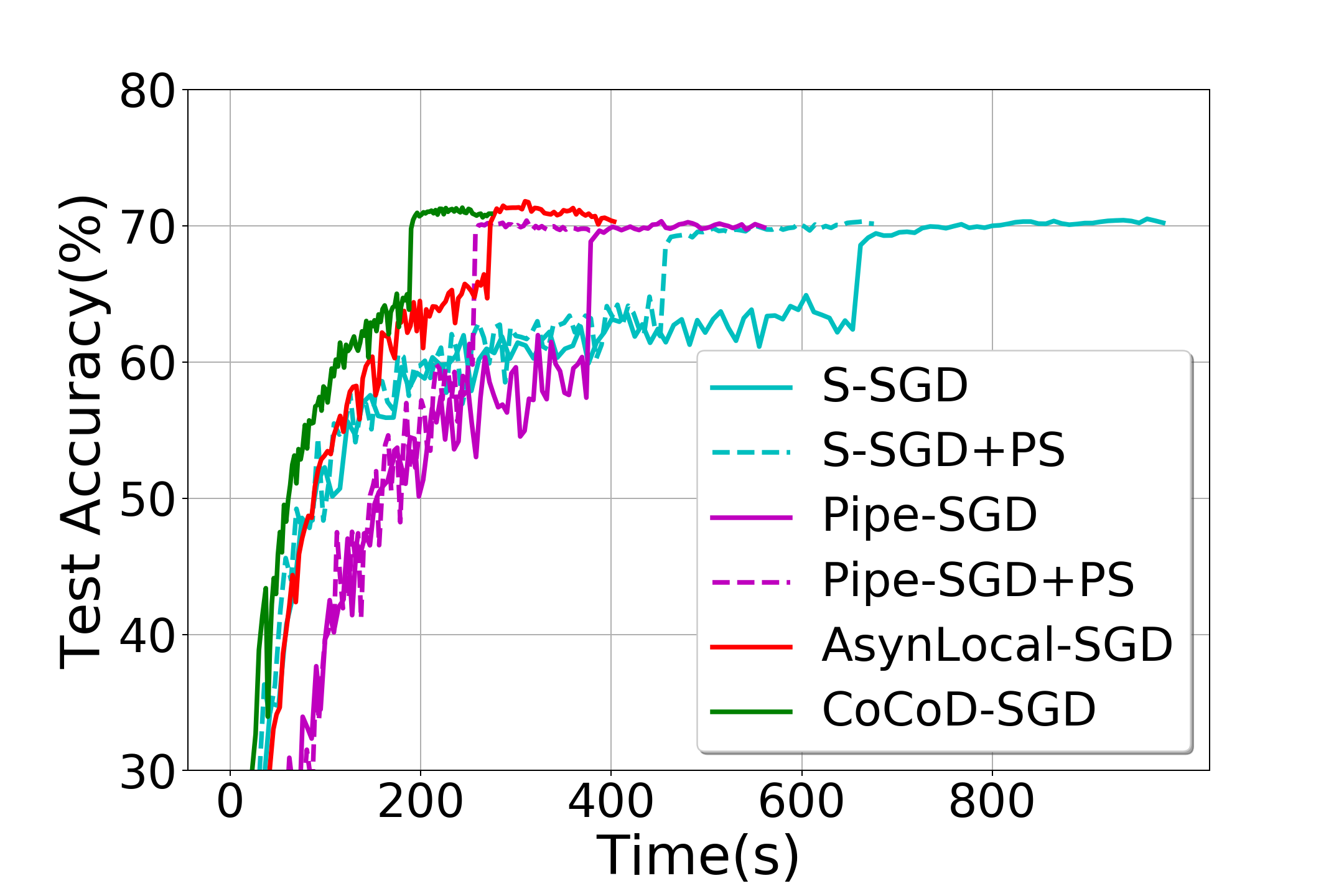}
\end{minipage}
}
\caption{Test accuracy for ResNet18 and VGG16 on CIFAR10 and CIFAR100 w.r.t time in a heterogeneous environment. CoCoD-SGD converges fastest and does not sacrifice the test accuracy.}
\label{Compare_time2_heter}
\end{figure*}

\subsection{Heterogeneous Environment}
To simulate a heterogeneous environment, we use 16 workers in our experiments and reduce the computation speed of 8 workers by half.
For S-SGD and Pipe-SGD, \emph{Proportionally Sampling} proposed in Section 3.2 can be also applied in the heterogeneous environment, which is 
equivalent to increasing the mini-batch size.
And we denote corresponding algorithms as S-SGD+PS and Pipe-SGD+PS respectively.
When one algorithm employs \emph{Proportionally Sampling}, the batch size is set to 32 for the slower workers and 64 for the faster workers.
In the heterogeneous environment with 16 workers, we set $\gamma_{16}^{\rm ps} = 0.01 * (8 + 8 * 2)$ for algorithms with \emph{Proportionally Sampling} and $\gamma_{16} = 0.01 * 16$ for others as suggested in Remark~\ref{remark_4}.
In addition, the asynchronous version of Local-SGD (AsynLocal-SGD), which is proposed for heterogeneous environments in~\citep{yu2018parallel}, is included 
in comparison with CoCoD-SGD 
in our experiments.
\paragraph{Comparison of convergence rate.} 
Figure~\ref{Compare_epoch_heter} presents the training loss regarding epochs of ResNet18 and VGG16 on 16 GPUs, which exhibits similar convergence rate for all the algorithms in comparison. Besides, we can see from Table~\ref{final_accuracy_heter} that CoCoD-SGD does not lose the test accuracy. 
\paragraph{Comparison of convergence speed.} 
Figure~\ref{Compare_time2_heter} shows the curves of the test accuracy regarding time. 
As we can see, CoCoD-SGD converges fastest and achieves almost 2.5$\times$ and 3$\times$ speedup against S-SGD for ResNet18 and VGG16 respectively. 
Comparing with the results in 
Figure~\ref{Compare_time},
 S-SGD converges slower due to the heterogeneous computation, while CoCoD-SGD is robust.
When equipped with \emph{Proportionally Sampling}, S-SGD and Pipe-SGD converge faster since the hardware utilization
is improved.

\begin{table}
\centering
\setlength{\tabcolsep}{1.25mm}{
\begin{tabular}{|l|l|l|l|l|}
\hline
& \multicolumn{2}{|c|}{CIFAR10} & \multicolumn{2}{|c|}{CIFAR100} \\
\hline
& ResNet18 & VGG16 & ResNet18 & VGG16 \\ 
\hline
S-SGD & 94.18\% & 93.09\% & 74.30\% & 70.51\% \\
\hline
S-SGD+PS & 93.73\% & 92.96\% & 74.16\% & 70.32\% \\
\hline
Pipe-SGD & 94.17\% & 92.36\% & 73.95\% & 70.26\%    \\
\hline
Pipe-SGD+PS & 94.10\% & 92.71\% & 73.84\% & 70.15\%    \\
\hline
AsynLocal-SGD & 94.47\% & 93.24\% & 75.21\% & 71.80\%    \\
\hline
CoCoD-SGD & 94.33\% & 92.69\% & 75.74\% & 71.33\%    \\
\hline
\end{tabular}}
\caption{Final best test accuracy for all tasks in a heterogeneous environment. 16 workers in total.}
\label{final_accuracy_heter}
\end{table}

\section{Conclusion}
In this paper, we propose a computation and communication decoupled stochastic gradient descent (CoCoD-SGD) for distributed optimization. 
In comparison with existing distributed algorithms, CoCoD-SGD can run computation and communication simultaneously to make full use of hardware resources and has lower communication complexity. 
CoCoD-SGD is also theoretically justified to have the same convergence rate as S-SGD and obtains linear iteration speedup in both homogeneous and heterogeneous environments. 
In addition, CoCoD-SGD achieves faster distributed training with superior time speedup when comparing with others.
Experimental results demonstrate the efficiency of the proposed algorithm.

\section*{Acknowledgments}
This research was supported by the National Natural Science Foundation of China (No. 61673364, No. 91746301) and the Fundamental Research Funds for the Central Universities (WK2150110008).
We also gratefully acknowledge the support of Cheng Li and Youhui Bai from USTC for providing the experimental environment.

\bibliographystyle{named}
\bibliography{arxiv-inline}

\newpage
\appendix

\begin{center}
{\Large \textbf{Supplemental Material for ``Faster Distributed Deep Net Training: Computation and Communication Decoupled Stochastic Gradient Descent''} }
\end{center}
\section*{Appendix: proofs}
At first, we bound the partially accumulated local gradients.
\begin{lemma} \label{bound_accu_grad}
Under Assumption 1, we have the following inequality
\begin{equation}
\sum_{i=1}^N \frac{M_i}{\sum_{l=1}^{N} M_l} \mathbb{E} \left \| \sum_{\tau = t'}^{t-1} \gamma G_{\tau}^i \right \|^2 \leq 4 \gamma^2 (t-t') \left( \frac{N\sigma^2}{\sum_{l=1}^{N} M_l} + \left(t - t'\right) \zeta^2 + L^2 \sum_{i=1}^N \frac{M_i}{\sum_{l=1}^{N} M_l} \sum_{\tau = t'}^{t-1} \mathbb{E} \left\| x_{\tau}^i - \hat{x}_{\tau} \right\|^2 +  \sum_{\tau = t'}^{t-1} \mathbb{E} \left\|  \nabla f(\hat{x}_{\tau}) \right\|^2 \right).
\end{equation}
\end{lemma}
\emph{Proof}. By the definition of $G_{\tau}^i$, we have
\begin{eqnarray} \label{lemma_bound_gradient}
&& \sum_{i=1}^N \frac{M_i}{\sum_{l=1}^{N} M_l} \mathbb{E} \left \| \sum_{\tau = t'}^{t-1} \gamma G_{\tau}^i \right \|^2
\nonumber\\
&=& \gamma^2 \sum_{i=1}^N \frac{M_i}{\sum_{l=1}^{N} M_l} \mathbb{E} \left \| \sum_{\tau = t'}^{t-1} \frac{1}{M_i} \sum_{j=1}^{M_i} \nabla f_{i}(x_{\tau}^i, \xi_{\tau}^{i,j}) \right \|^2
\nonumber\\
&\leq& 4 \gamma^2 \sum_{i=1}^N \frac{M_i}{\sum_{l=1}^{N} M_l} \left ( \underbrace{\mathbb{E} \left \| \sum_{\tau = t'}^{t-1} \left( \frac{1}{M_i} \sum_{j=1}^{M_i}  \nabla f_{i}\left(x_{\tau}^i, \xi_{\tau}^{i,j}\right) - \nabla f_{i}\left(x_{\tau}^i\right) \right) \right \|^2}_{T_1} + \underbrace{\mathbb{E} \left \| \sum_{\tau = t'}^{t-1}\left( \nabla f_{i}\left(x_{\tau}^i\right) - \nabla f_{i}\left(\hat{x}_{\tau}\right) \right) \right\|^2}_{T_2} \right.\
\nonumber\\
&& \left.\ + \underbrace{\mathbb{E} \left\| \sum_{\tau = t'}^{t-1}\left( \nabla f_i\left(\hat{x}_{\tau}\right) - \nabla f\left(\hat{x}_{\tau}\right)\right) \right\|^2}_{T_3} + \underbrace{\mathbb{E} \left\| \sum_{\tau = t'}^{t-1} \nabla f\left(\hat{x}_{\tau}\right) \right\|^2}_{T_4} \right ),
\end{eqnarray}
where the inequality follows from Cauchy's inequality. We next bound $T_1$
\begin{eqnarray} \label{lemma_bound_T_1}
T_1 &=& \sum_{\tau = t'}^{t-1} \mathbb{E} \left \| \frac{1}{M_i} \sum_{j=1}^{M_i}  \nabla f_{i}\left(x_{\tau}^i, \xi_{\tau}^{i,j}\right) - \nabla f_{i}\left(x_{\tau}^i\right) \right \|^2 
\nonumber\\
&&+ 2 \sum_{t' \leq \tau_1 < \tau_2 \leq t-1} \mathbb{E} \left \langle \frac{1}{M_i} \sum_{j=1}^{M_i}  \nabla f_{i}\left(x_{\tau_1}^i, \xi_{\tau_1}^{i,j}\right) - \nabla f_{i}\left(x_{\tau_1}^i\right), \frac{1}{M_i} \sum_{j=1}^{M_i}  \nabla f_{i}\left(x_{\tau_2}^i, \xi_{\tau_2}^{i,j}\right) - \nabla f_{i}\left(x_{\tau_2}^i\right)  \right \rangle
\nonumber\\
&=&  \sum_{\tau = t'}^{t-1} \mathbb{E} \left \| \frac{1}{M_i} \sum_{j=1}^{M_i}  \nabla f_{i}\left(x_{\tau}^i, \xi_{\tau}^{i,j}\right) - \nabla f_{i}\left(x_{\tau}^i\right) \right \|^2
\nonumber\\
&=& \sum_{\tau = t'}^{t-1}  \frac{1}{M_i^2} \left( \sum_{j=1}^{M_i} \mathbb{E} \left \| \nabla f_{i}(x_{\tau}^i, \xi_{\tau}^{i,j}) - \nabla f_{i}(x_{\tau}^i) \right\|^2 \right.
\nonumber\\
&&\left. + 2\sum_{1\leq j_1 < j_2  \leq M_i} \mathbb{E} \left \langle \nabla f_{i}(x_{\tau}^i, \xi_{\tau}^{i,j_1}) - \nabla f_{i}(x_{\tau}^i), \nabla f_{i}(x_{\tau}^i, \xi_{\tau}^{i,j_2}) - \nabla f_{i}(x_{\tau}^i) \right \rangle \right)
\nonumber\\
&=& \sum_{\tau = t'}^{t-1}  \frac{1}{M_i^2} \sum_{j=1}^{M_i} \mathbb{E} \left \|    \nabla f_{i}(x_{\tau}^i, \xi_{\tau}^{i,j}) - \nabla f_{i}(x_{\tau}^i) \right\|^2
\nonumber\\
&\leq& \frac{\left(t - t'\right) \sigma^2}{M_i},
\end{eqnarray}
where the second and the fourth equalities hold because
$\mathbb{E}_{\xi_{\tau}^{i,j} \in \mathcal{D}_{i}} \nabla f_{i}(x_{\tau}^i, \xi_{\tau}^{i,j}) = \nabla f_{i}(x_{\tau}^i)$
and $\xi_{\tau}^{i,j}$'s are independent, and the inequality follows from Assumption~1~(3). 
According to Cauchy's inequality, we can bound $T_2$, $T_3$ and $T_4$ as
\begin{eqnarray}
T_2 &\leq& \left(t - t'\right) \sum_{\tau = t'}^{t-1} \left \|  \nabla f_{i}\left(x_{\tau}^i\right) - \nabla f_{i}\left(\hat{x}_{\tau}\right) \right \|^2 \leq \left(t - t'\right) L^2 \sum_{\tau = t'}^{t-1} \left \|  x_{\tau}^i - \hat{x}_{\tau} \right \|^2, \label{lemma_bound_T_2}
\\
\sum_{i=1}^N \frac{M_i}{\sum_{l=1}^{N} M_l} T_3 &\leq& \left(t - t'\right)\sum_{\tau = t'}^{t-1} \mathbb{E} \sum_{i=1}^N \frac{M_i}{\sum_{l=1}^{N} M_l} \left\| \nabla f_i\left(\hat{x}_{\tau}\right) - \nabla f\left(\hat{x}_{\tau}\right) \right\|^2 \leq \left(t-t'\right)^2 \zeta^2, \label{lemma_bound_T_3}
\\
T_4 &\leq& (t - t') \sum_{\tau = t'}^{t-1} \mathbb{E} \left\| \nabla f(\hat{x}_{\tau}) \right\|^2,\label{lemma_bound_T_4}
\end{eqnarray}
where the second inequality in (\ref{lemma_bound_T_2}) and the second inequality in (\ref{lemma_bound_T_3}) follow Assumption~1 (1) and (3), respectively. Substituting (\ref{lemma_bound_T_1}), (\ref{lemma_bound_T_2}), (\ref{lemma_bound_T_3}) and (\ref{lemma_bound_T_4}) into (\ref{lemma_bound_gradient}), we obtain
\begin{eqnarray}
&&\sum_{i=1}^N \frac{M_i}{\sum_{l=1}^{N} M_l} \mathbb{E} \left \| \sum_{\tau = t'}^{t-1} \gamma G_{\tau}^i \right \|^2
\nonumber\\
&\leq& 4 \gamma^2  \left( \frac{N\left(t - t'\right) \sigma^2}{\sum_{l=1}^{N} M_l} + \left(t - t'\right) L^2 \sum_{i=1}^N \frac{M_i}{\sum_{l=1}^{N} M_l} \sum_{\tau = t'}^{t-1} \left \|  x_{\tau}^i - \hat{x}_{\tau} \right \|^2 + \left(t-t'\right)^2 \zeta^2 + (t - t') \sum_{\tau = t'}^{t-1} \mathbb{E} \left\| \nabla f(\hat{x}_{\tau}) \right\|^2 \right)
\nonumber\\
&=& 4 \gamma^2 (t-t') \left( \frac{N\sigma^2}{\sum_{l=1}^{N} M_l} + \left(t - t'\right) \zeta^2 + L^2 \sum_{i=1}^N \frac{M_i}{\sum_{l=1}^{N} M_l} \sum_{\tau = t'}^{t-1} \mathbb{E} \left\| x_{\tau}^i - \hat{x}_{\tau} \right\|^2 +  \sum_{\tau = t'}^{t-1} \mathbb{E} \left\|  \nabla f(\hat{x}_{\tau}) \right\|^2 \right),
\end{eqnarray}
which completes the proof.

Next, we bound the difference between the local models and the global average model.
\begin{lemma} \label{bound_difference}
Under Assumption~1, the difference of $\hat{x}_{t}$ and $x_{t}^i$'s can be bounded as
\begin{equation}\label{bound_difference_inequality}
\sum_{i=1}^{N} \frac{M_i}{\sum_{l=1}^{N} M_l} \sum_{t=0}^{T-1} \mathbb{E} \left\| \hat{x}_t - x_t^i \right\|^2
\leq \frac{8 \gamma^2 k}{1 - 16 \gamma^2 k^2 L^2} \left( \frac{TN\sigma^2}{\sum_{l=1}^{N} M_l} + 2kT\zeta^2 + 2k\sum_{t=0}^{T-1}  \mathbb{E} \left\|  \nabla f(\hat{x}_{t}) \right\|^2 \right).
\end{equation}
\end{lemma}
\emph{Proof.} 
According to the updating scheme in Algorithms~1, $x_t^i$ can be represented as
\begin{equation}\label{local_represent}
x_{t}^{i} = \hat{x}_{(\lfloor \frac{t}{k} \rfloor - 1) k} - \sum_{\tau = (\lfloor \frac{t}{k} \rfloor -1) k}^{t-1} \gamma G_{\tau}^{i},
\end{equation}
since the result of the last complete communication is the average of the models at step $(\lfloor \frac{t}{k} \rfloor - 1) k$. 
On the other hand, by the definition of $\hat{x}_t$, we can represent it as
\begin{equation}\label{average_represent}
\hat{x}_t = \hat{x}_{(\lfloor \frac{t}{k} \rfloor - 1) k} - \sum_{\tau = (\lfloor \frac{t}{k} \rfloor - 1) k}^{t-1} \gamma  \sum_{j=1}^N \frac{M_j}{\sum_{l=1}^{N} M_l} G_{\tau}^j.
\end{equation}
Substituting (\ref{local_represent}) and (\ref{average_represent}) into the left hand side of (\ref{bound_difference_inequality}) , we have
\begin{eqnarray}
&&\sum_{i=1}^{N} \frac{M_i}{\sum_{l=1}^{N} M_l} \mathbb{E} \left\| \hat{x}_t - x_t^i \right\|^2
\nonumber\\
&=& \sum_{i=1}^{N} \frac{M_i}{\sum_{l=1}^{N} M_l} \mathbb{E} \left\| \left(\hat{x}_{t'} - \sum_{\tau = t'}^{t-1} \gamma  \sum_{j=1}^N \frac{M_j}{\sum_{l=1}^{N} M_l} G_{\tau}^j\right) - \left(\hat{x}_{t'} - \sum_{\tau = t'}^{t-1} \gamma G_{\tau}^i\right)  \right\|^2
\nonumber\\
&=& \sum_{i=1}^{N} \frac{M_i}{\sum_{l=1}^{N} M_l} \mathbb{E} \left\| \sum_{\tau = t'}^{t-1} \gamma G_{\tau}^i- \sum_{\tau = t'}^{t-1} \gamma \sum_{j=1}^N \frac{M_j}{\sum_{l=1}^{N} M_l} G_{\tau}^j \right\|^2
\nonumber\\
&=& \sum_{i=1}^{N} \frac{M_i}{\sum_{l=1}^{N} M_l} \mathbb{E} \left\| \sum_{\tau = t'}^{t-1} \gamma G_{\tau}^i\right\|^2 + \sum_{i=1}^{N} \frac{M_i}{\sum_{l=1}^{N} M_l} \mathbb{E} \left\| \sum_{\tau = t'}^{t-1} \gamma \sum_{j=1}^N \frac{M_j}{\sum_{l=1}^{N} M_l} G_{\tau}^j \right\|^2  
\nonumber\\
&&- 2\sum_{i=1}^{N} \frac{M_i}{\sum_{l=1}^{N} M_l} \mathbb{E} \left\langle\sum_{\tau = t'}^{t-1} \gamma G_{\tau}^i,  \sum_{\tau = t'}^{t-1} \gamma \sum_{j=1}^N \frac{M_j}{\sum_{l=1}^{N} M_l} G_{\tau}^j \right\rangle
\nonumber\\
&=& \frac{1}{N} \sum_{i=1}^{N} \mathbb{E} \| \sum_{\tau = t'}^{t-1} \gamma G_{\tau}^i\|^2 +  \mathbb{E} \| \sum_{\tau = t'}^{t-1} \gamma \sum_{j=1}^N \frac{M_j}{\sum_{l=1}^{N} M_l} G_{\tau}^j \|^2 - 2\mathbb{E} \| \sum_{\tau = t'}^{t-1} \gamma \sum_{j=1}^N \frac{M_j}{\sum_{l=1}^{N} M_l} G_{\tau}^j \|^2 
\nonumber\\
&=& \sum_{i=1}^{N} \frac{M_i}{\sum_{l=1}^{N} M_l} \mathbb{E} \left\| \sum_{\tau = t'}^{t-1} \gamma G_{\tau}^i \right\|^2  -  \mathbb{E} \left\| \sum_{\tau = t'}^{t-1} \gamma \sum_{j=1}^N \frac{M_j}{\sum_{l=1}^{N} M_l} G_{\tau}^j \right\|^2 
\nonumber\\
&\leq& \sum_{i=1}^{N} \frac{M_i}{\sum_{l=1}^{N} M_l} \mathbb{E} \left\| \sum_{\tau = t'}^{t-1} \gamma G_{\tau}^i \right\|^2 
\nonumber\\
&\leq& 4 \gamma^2 (t-t') \left( \frac{N\sigma^2}{\sum_{l=1}^{N} M_l} + \left(t - t'\right) \zeta^2 + L^2 \sum_{i=1}^N \frac{M_i}{\sum_{l=1}^{N} M_l} \sum_{\tau = t'}^{t-1} \mathbb{E} \left\| x_{\tau}^i - \hat{x}_{\tau} \right\|^2 +  \sum_{\tau = t'}^{t-1} \mathbb{E} \left\|  \nabla f(\hat{x}_{\tau}) \right\|^2 \right),
\end{eqnarray}
where the last inequality follows from Lemma~\ref{bound_accu_grad}. Since $t' = (\lfloor \frac{t}{k} \rfloor - 1)k$, we have $t' \geq t- 2k$ and can further obtain
\begin{eqnarray}
&&\sum_{i=1}^{N} \frac{M_i}{\sum_{l=1}^{N} M_l} \mathbb{E} \| \hat{x}_t - x_t^i \|^2
\nonumber\\
&\leq& 8 \gamma^2 k \left ( \frac{N\sigma^2}{\sum_{l=1}^{N} M_l} + 2k\zeta^2 + L^2 \sum_{i=1}^N \frac{M_i}{\sum_{l=1}^{N} M_l} \sum_{\tau = t-2k}^{t-1} \mathbb{E} \| x_{\tau}^i - \hat{x}_{\tau}\|^2 +  \sum_{\tau = t-2k}^{t-1} \mathbb{E} \|  \nabla f(\hat{x}_{\tau}) \|^2 \right ).
\end{eqnarray}
Summing up this inequality from $t=0$ to $T-1$, we have
\begin{eqnarray}
&&\sum_{i=1}^{N} \frac{M_i}{\sum_{l=1}^{N} M_l} \sum_{t=0}^{T-1} \mathbb{E} \| \hat{x}_t - x_t^i \|^2
\nonumber\\
&\leq& 8 \gamma^2 k \left( \frac{TN\sigma^2}{\sum_{l=1}^{N} M_l} + 2kT\zeta^2 + L^2 \sum_{i=1}^{N} \frac{M_i}{\sum_{l=1}^{N} M_l} \sum_{t=0}^{T-1} \sum_{\tau = t-2k}^{t-1} \mathbb{E} \left\| x_{\tau}^i - \hat{x}_{\tau}\right\|^2 +  \sum_{t=0}^{T-1} \sum_{\tau = t-2k}^{t-1} \mathbb{E} \left\|  \nabla f\left(\hat{x}_{\tau}\right) \right\|^2 \right )
\nonumber\\
&\leq& 8 \gamma^2 k \left( \frac{TN\sigma^2}{\sum_{l=1}^{N} M_l} + 2kT\zeta^2 + 2k L^2 \sum_{i=1}^{N} \frac{M_i}{\sum_{l=1}^{N} M_l} \sum_{t=0}^{T-1}  \mathbb{E} \left\| x_{t}^i - \hat{x}_{t} \right\|^2 +  2k\sum_{t=0}^{T-1}  \mathbb{E} \left\|  \nabla f(\hat{x}_{t}) \right\|^2 \right),
\end{eqnarray}
where the last inequality can be obtained by using a simple counting argument $\sum_{t=0}^{T-1} \sum_{\tau = t-2k}^{t-1} A_{\tau} \leq 2k \sum_{t=0}^{T-1} A_t$. Rearranging the inequality, we obtain
\begin{eqnarray}
\left(1 - 16 \gamma^2 k^2 L^2\right) \sum_{i=1}^{N} \frac{M_i}{\sum_{l=1}^{N} M_l} \sum_{t=0}^{T-1} \mathbb{E} \left\| \hat{x}_t - x_t^i \right\|^2
\leq 8 \gamma^2 k  \left( \frac{TN\sigma^2}{\sum_{l=1}^{N} M_l} + 2kT\zeta^2 + 2k\sum_{t=0}^{T-1}  \mathbb{E} \left\|  \nabla f(\hat{x}_{t}) \right\|^2 \right).
\end{eqnarray}
Dividing $\left(1 - 16 \gamma^2 k^2 L^2\right)$ on both sides yields the result.

\begin{theorem} \label{general_theorem_appendix}
Under Assumption~1, if the learning rate satisfies $\gamma \leq \frac{1}{L}$, we have the following convergence result for Algorithm~1:
\begin{eqnarray} \label{general_threorem_result_appendix}
\frac{1}{T} \sum_{t=0}^{T-1} D_1 \mathbb{E} \| \nabla f(\hat{x}_t) \|^2
\leq\frac{ 2(f(\hat{x}_0) - f^*)}{T\gamma} +  D_2 (\frac{N\sigma^2}{\sum_{i=1}^{N} M_i} + 2k\zeta^2) + \frac{ \gamma L \sigma^2}{\sum_{i=1}^{N} M_i},~~~~
\end{eqnarray}
where
\begin{equation}
D_1 = 1 - 2kD_2, ~~~~D_2 = \frac{8 \gamma^2 L^2 k}{1 - 16 \gamma^2 k^2 L^2}.
\end{equation}
\end{theorem}
\emph{Proof.} Since $f_i(\cdot), i=1, 2, \cdots, N$ are $L$-smooth, it is easy to verify that $f(\cdot)$ is $L$-smooth. We have
\begin{eqnarray}
f(\hat{x}_{t+1}) &\leq& f(\hat{x}_t) + \left \langle \nabla f(\hat{x}_t), \hat{x}_{t+1} - \hat{x}_t \right\rangle + \frac{L}{2} \left \| \hat{x}_{t+1} - \hat{x}_t \right \|^2
\nonumber\\
&=& f(\hat{x}_t) - \gamma \left \langle  \nabla f(\hat{x}_t), \sum_{i=1}^N \frac{M_i}{\sum_{l=1}^{N} M_l} G_i^t  \right \rangle + \frac{L \gamma^2}{2} \left \| \sum_{i=1}^N \frac{M_i}{\sum_{l=1}^{N} M_l} G_t^i \right \|^2.
\end{eqnarray}
By applying expectation with respect to all the random variables at step $t$ and conditional on the past (denote by $\mathbb{E}_{t|\cdot}$), we have
\begin{eqnarray} \label{nonconvex_bound_exp}
&&\mathbb{E}_{t|\cdot} f(\hat{x}_{t+1}) 
\nonumber\\
&\leq& f(\hat{x}_t) - \gamma \left \langle \nabla f(\hat{x}_t), \sum_{i=1}^N \frac{M_i}{\sum_{l=1}^{N} M_l} \nabla f_i (x_t^i) \right \rangle + \frac{L \gamma^2}{2} \mathbb{E}_{t|\cdot} \left \| \sum_{i=1}^N \frac{M_i}{\sum_{l=1}^{N} M_l} G_t^i \right\|^2
\nonumber\\
&=& f(\hat{x}_t) - \frac{\gamma}{2} \left ( \left\| \nabla f(\hat{x}_t) \right\|^2 + \left \| \sum_{i=1}^N \frac{M_i}{\sum_{l=1}^{N} M_l} \nabla f_i (x_t^i) \right\|^2 - \left \| \nabla f(\hat{x}_t) - \sum_{i=1}^N \frac{M_i}{\sum_{l=1}^{N} M_l} \nabla f_i (x_t^i) \right \|^2 \right) 
\nonumber\\
&&+ \frac{L \gamma^2}{2} \mathbb{E}_{t|\cdot} \left \| \sum_{i=1}^N \frac{M_i}{\sum_{l=1}^{N} M_l} G_t^i \right\|^2.
\end{eqnarray}
Note that
\begin{eqnarray}
&&\mathbb{E}_{t|\cdot} \left \| \sum_{i=1}^N \frac{M_i}{\sum_{l=1}^{N} M_l} G_t^i \right\|^2
\nonumber\\
&=& \mathbb{E}_{t|\cdot} \left\| \sum_{i=1}^N \frac{M_i}{\sum_{l=1}^{N} M_l} G_t^i - \sum_{i=1}^N \frac{M_i}{\sum_{l=1}^{N} M_l} \nabla f_i (x_t^i) + \sum_{i=1}^N \frac{M_i}{\sum_{l=1}^{N} M_l} \nabla f_i (x_t^i) \right\|^2
\nonumber\\
&=& \mathbb{E}_{t|\cdot} \left \| \sum_{i=1}^N \frac{M_i}{\sum_{l=1}^{N} M_l} G_t^i - \sum_{i=1}^N \frac{M_i}{\sum_{l=1}^{N} M_l} \nabla f_i (x_t^i) \right\|^2 +  \left\| \sum_{i=1}^N \frac{M_i}{\sum_{l=1}^{N} M_l} \nabla f_i (x_t^i) \right\|^2 
\nonumber\\
&&+ 2 \mathbb{E}_{t|\cdot} \left \langle \sum_{i=1}^N \frac{M_i}{\sum_{l=1}^{N} M_l} G_t^i - \sum_{i=1}^N \frac{M_i}{\sum_{l=1}^{N} M_l} \nabla f_i (x_t^i), \sum_{i=1}^N \frac{M_i}{\sum_{l=1}^{N} M_l} \nabla f_i (x_t^i) \right \rangle
\nonumber\\
&=&  \mathbb{E}_{t|\cdot} \left \| \sum_{i=1}^N \frac{M_i}{\sum_{l=1}^{N} M_l} G_t^i - \sum_{i=1}^N \frac{M_i}{\sum_{l=1}^{N} M_l} \nabla f_i (x_t^i) \right\|^2 +  \left \| \sum_{i=1}^N \frac{M_i}{\sum_{l=1}^{N} M_l} \nabla f_i (x_t^i) \right\|^2,
\end{eqnarray}
where the last equality holds because $ \mathbb{E}_{t|\cdot} \left( \frac{1}{N} \sum_{i=1}^N G_t^i - \frac{1}{N} \sum_{i=1}^N \nabla f_i (x_i^t) \right) = 0$, and
\begin{eqnarray}
&&\mathbb{E}_{t|\cdot} \left \| \sum_{i=1}^N \frac{M_i}{\sum_{l=1}^{N} M_l} G_t^i - \sum_{i=1}^N \frac{M_i}{\sum_{l=1}^{N} M_l} \nabla f_i (x_t^i) \right\|^2
\nonumber\\
&=& \mathbb{E}_{t|\cdot} \sum_{i=1}^N \frac{M_i^2}{(\sum_{l=1}^{N} M_l)^2} \left \| G_t^i - \nabla f_i (x_t^i) \right\|^2 
\nonumber\\
&&+ 2 \sum_{1\leq i_1 < i_2 \leq N} \mathbb{E}_{t|\cdot} \left \langle \frac{M_{i_1}}{\sum_{l=1}^{N} M_l} G_t^{i_1} - \frac{M_{i_1}}{\sum_{l=1}^{N} M_l} \nabla f_{i_1} (x_t^{i_1}),  \frac{M_{i_2}}{\sum_{l=1}^{N} M_l} G_t^{i_2} - \frac{M_{i_2}}{\sum_{l=1}^{N} M_l} \nabla f_{i_2} (x_t^{i_2}) \right \rangle
\nonumber\\
&=& \mathbb{E}_{t|\cdot} \sum_{i=1}^N \frac{M_i^2}{(\sum_{l=1}^{N} M_l)^2} \left \| G_t^i - \nabla f_i (x_t^i) \right\|^2
\nonumber\\
&=& \mathbb{E}_{t|\cdot} \sum_{i=1}^N \frac{M_i^2}{(\sum_{l=1}^{N} M_l)^2} \left \| \frac{1}{M_i} \sum_{j=1}^{M_i} \nabla f_i(x_t^i, \xi_{t}^{i,j}) - \nabla f_i (x_t^i) \right \|^2
\nonumber\\
&=& \mathbb{E}_{t|\cdot} \sum_{i=1}^N \frac{1}{(\sum_{l=1}^{N} M_l)^2} \Bigg ( \sum_{j=1}^{M_i} \left \| \nabla f_i(x_t^i, \xi_{t}^{i,j}) - \nabla f_i (x_t^i) \right\|^2 
\nonumber\\
&&+ 2 \sum_{1\leq j_1 <  j_2 \leq M_i} \left \langle \nabla f_i(x_t^i, \xi_{t}^{i,j_1}) - \nabla f_i (x_t^i), \nabla f_i(x_t^i, \xi_{t}^{i,j_2}) - \nabla f_i (x_t^i)  \right \rangle \Bigg)
\nonumber\\
&=& \sum_{i=1}^N \frac{1}{(\sum_{l=1}^{N} M_l)^2} \sum_{j=1}^{M_i} \mathbb{E}_{t|\cdot} \| \nabla f_i(x_t^i, \xi_{t}^{i,j}) - \nabla f_i (x_t^i) \|^2
\nonumber\\
&\leq& \sum_{i=1}^N \frac{1}{(\sum_{l=1}^{N} M_l)^2} M_i \sigma^2 = \frac{\sigma^2}{\sum_{l=1}^{N} M_l},
\end{eqnarray}
where the second equality and the fifth equality hold because the random variables on different workers and the random variables in one mini-batch are independent, and the last inequality follows from Assumption~1~(3). We have
\begin{equation} \label{nonconvex_bound_second}
\mathbb{E}_{t|\cdot} \left\| \sum_{i=1}^N \frac{M_i}{\sum_{l=1}^{N} M_l} G_i^t \right\|^2 \leq \frac{\sigma^2}{\sum_{l=1}^{N} M_l} + \mathbb{E}_{t|\cdot} \left\| \sum_{i=1}^N \frac{M_i}{\sum_{l=1}^{N} M_l} \nabla f_i (x_i^t)\right\|^2.
\end{equation}
Substituting (\ref{nonconvex_bound_second}) into (\ref{nonconvex_bound_exp}) and applying expectation with respect to all the random variables, we obtain
\begin{eqnarray} \label{nonconvex_bound_F}
\mathbb{E} f(\hat{x}_{t+1}) &\leq& \mathbb{E} f(\hat{x}_t) - \frac{\gamma}{2} \mathbb{E} \| \nabla f(\hat{x}_t) \|^2 - \frac{\gamma}{2} (1 - L \gamma) \mathbb{E} \left \| \sum_{i=1}^N \frac{M_i}{\sum_{l=1}^{N} M_l} \nabla f_i (x_t^i) \right \|^2 
\nonumber\\
&& + \frac{\gamma}{2} \mathbb{E} \left\| \nabla f(\hat{x}_t) - \sum_{i=1}^N \frac{M_i}{\sum_{l=1}^{N} M_l} \nabla f_i (x_t^i) \right\|^2 + \frac{\gamma^2 L \sigma^2}{2\sum_{l=1}^{N} M_l}.
\end{eqnarray}
We then bound the difference of $\nabla f(\hat{x}^t)$ and $\sum_{i=1}^N \frac{M_i}{\sum_{l=1}^{N} M_l} \nabla f_i (x_i^t)$ as
\begin{eqnarray} \label{nonconvex_bound_grad_difference}
\mathbb{E} \left\| \nabla f(\hat{x}^t) - \sum_{i=1}^N \frac{M_i}{\sum_{l=1}^{N} M_l} \nabla f_i (x_i^t) \right\|^2 & =& \mathbb{E} \left\| \sum_{i=1}^N \frac{M_i}{\sum_{l=1}^{N} M_l} \left( \nabla f_i(\hat{x}^t) - \nabla f_i(x_i^t) \right) \right\|^2
\nonumber\\
&\leq& \mathbb{E} \left(\sum_{i=1}^{N} \left(\frac{\sqrt{M_i}}{\sum_{l=1}^{N} M_l}\right)^2\right) \left(\sum_{i=1}^N  \left\| \sqrt{M_i} \left(\nabla f_i(\hat{x}^t) - \nabla f_i(x_i^t) \right) \right\|^2 \right)
\nonumber\\
&=& \sum_{i=1}^N  \frac{M_i}{\sum_{l=1}^{N} M_l} \mathbb{E} \left\| \nabla f_i(\hat{x}^t) - \nabla f_i(x_i^t) \right\|^2
\nonumber\\
&\leq& L^2 \sum_{i=1}^N \frac{M_i}{\sum_{l=1}^{N} M_l} \mathbb{E} \left\| \hat{x}^t - x_i^t \right\|^2,
\end{eqnarray}
where the two inequalities follow from Cauchy's inequality and $L$-smooth assumption, respectively. Substituting (\ref{nonconvex_bound_grad_difference}) into (\ref{nonconvex_bound_F}) yields
\begin{eqnarray}
\mathbb{E} f(\hat{x}_{t+1}) \leq \mathbb{E} f(\hat{x}_t) - \frac{\gamma}{2} \mathbb{E} \| \nabla f(\hat{x}_t) \|^2 - \frac{\gamma}{2} (1 - L \gamma) \mathbb{E} \left\| \sum_{i=1}^N \frac{M_i}{\sum_{l=1}^{N} M_l} \nabla f_i (x_t^i) \right\|^2 
\nonumber\\
 + \frac{\gamma L^2}{2} \sum_{i=1}^N \frac{M_i}{\sum_{l=1}^{N} M_l} \mathbb{E} \| \tilde{x}^t - x_i^t \|^2 + \frac{\gamma^2 L \sigma^2}{2\sum_{l=1}^{N} M_l}.
\end{eqnarray}
Rearranging the inequality and summing up both sides from $t=0$ to $T-1$, we have
\begin{eqnarray}\label{bound_sum_1}
&&\sum_{t=0}^{T-1} \left(\frac{\gamma}{2} \mathbb{E} \| \nabla f(\hat{x}_t) \|^2 + \frac{\gamma}{2} (1 - L \gamma) \mathbb{E} \left\| \sum_{i=1}^N \frac{M_i}{\sum_{l=1}^{N} M_l} \nabla f_i (x_t^i) \right\|^2 \right)
\nonumber\\
&\leq& f(\hat{x}_0) - f^* +  \frac{\gamma L^2}{2} \sum_{i=1}^N \frac{M_i}{\sum_{l=1}^{N} M_l} \sum_{t=0}^{T-1} \mathbb{E} \| \tilde{x}^t - x_i^t \|^2 + \frac{T \gamma^2 L \sigma^2}{2\sum_{l=1}^{N} M_l}.
\end{eqnarray}
Substituting Lemma~\ref{bound_difference} into (\ref{bound_sum_1}), we obtain
\begin{eqnarray}\label{bound_sum_2}
&&\sum_{t=0}^{T-1} \left(\frac{\gamma}{2} \mathbb{E} \| \nabla f(\hat{x}_t) \|^2 + \frac{\gamma}{2} (1 - L \gamma) \mathbb{E} \left\| \sum_{i=1}^N \frac{M_i}{\sum_{l=1}^{N} M_l} \nabla f_i (x_t^i) \right\|^2 \right)
\nonumber\\
&\leq& f(\hat{x}_0) - f^* +  \frac{4 \gamma^3 L^2 k}{1 - 16 \gamma^2 k^2 L^2} \left( T\left(\frac{N\sigma^2}{\sum_{l=1}^{N} M_l} + 2k \zeta^2\right) + 2k\sum_{t=0}^{T-1}  \mathbb{E} \|  \nabla f(\hat{x}_{t}) \|^2 \right) + \frac{T \gamma^2 L \sigma^2}{2\sum_{l=1}^{N} M_l}.
\end{eqnarray}
Rearranging this inequality and dividing both sides by $\frac{T \gamma}{2}$, we get
\begin{eqnarray}\label{bound_sum_3}
&&\frac{1}{T} \sum_{t=0}^{T-1} \left( \left(1 - \frac{16\gamma^2 L^2 k^2}{1 - 16 \gamma^2 k^2 L^2} \right)\mathbb{E} \| \nabla f(\hat{x}_t) \|^2 + (1 - L \gamma)  \mathbb{E} \left\| \sum_{i=1}^N \frac{M_i}{\sum_{l=1}^{N} M_l} \nabla f_i (x_t^i) \right\|^2 \right)
\nonumber\\
&\leq& \frac{ 2(f(\hat{x}_0) - f(\hat{x}_T))}{T\gamma} +  \frac{8 \gamma^2 L^2 k}{1 - 16 \gamma^2 k^2 L^2} \left(\frac{N\sigma^2}{\sum_{l=1}^{N} M_l} + 2k\zeta^2\right) + \frac{ \gamma L \sigma^2}{\sum_{l=1}^{N} M_l}
\nonumber\\
&\leq& \frac{ 2(f(\hat{x}_0) - f^*)}{T\gamma} +  \frac{8 \gamma^2 L^2 k}{1 - 16 \gamma^2 k^2 L^2} \left(\frac{N\sigma^2}{\sum_{l=1}^{N} M_l} + 2k\zeta^2\right) + \frac{ \gamma L \sigma^2}{\sum_{l=1}^{N} M_l}.
\end{eqnarray}
If the learning rate satisfies $\gamma \leq \frac{1}{L}$, we have
\begin{eqnarray}\label{bound_sum_4}
\frac{1}{T} \sum_{t=0}^{T-1} \left(1 - \frac{16\gamma^2 L^2 k^2}{1 - 16 \gamma^2 k^2 L^2} \right)\mathbb{E} \| \nabla f(\hat{x}_t) \|^2
\leq\frac{ 2(f(\hat{x}_0) - f^*)}{T\gamma} +  \frac{8 \gamma^2 L^2 k}{1 - 16 \gamma^2 k^2 L^2} \left(\frac{N\sigma^2}{\sum_{l=1}^{N} M_l} + 2k\zeta^2\right) + \frac{ \gamma L \sigma^2}{\sum_{l=1}^{N} M_l},
\end{eqnarray}
which completes the proof.

\begin{corollary} \label{general_corollary_appendix}
Under Assumption~1,
when the learning rate is set as $\gamma = \frac{1}{\sigma \sqrt{\frac{T}{\sum_{i=1}^{N} M_i}}}$ and the total number of iterations satisfies
{\small
\begin{eqnarray}
T \geq \max \Bigg \{ \frac{L^2(\sum_{i=1}^{N} M_i)}{\sigma^2}, \frac{48(\sum_{i=1}^{N} M_i)L^2k^2}{\sigma^2},  
\frac{144(\sum_{i=1}^{N} M_i)^3}{\sigma^6}L^2k^2 \left( \frac{N\sigma^2}{\sum_{i=1}^{N} M_i} + 2k\zeta^2 \right)^2 \Bigg\},~~ 
\end{eqnarray}
}
we have the following convergence result for Algorithm~1:
\begin{equation}
\frac{1}{T} \sum_{t=0}^{T-1} \mathbb{E} \| \nabla f(\hat{x}_t) \|^2 \leq \frac{4\sigma(f(\hat{x}_0) - f^* + L)}{\sqrt{T\sum_{i=1}^{N} M_i}}.
\end{equation}
\end{corollary}
\emph{Proof.} 
Since $\gamma = \frac{1}{\sigma \sqrt{\frac{T}{\sum_{l=1}^{N} M_l}}}$ and $T \geq \frac{L^2\sum_{l=1}^{N} M_l}{\sigma^2}$, we immediately have $\gamma \leq \frac{1}{L}$, then we have the result in (\ref{general_threorem_result_appendix}) and get
\begin{equation}\label{corollary_convergence}
\frac{1}{T} \sum_{t=0}^{T-1} \mathbb{E} \| \nabla f(\hat{x}_t) \|^2 
\leq \frac{ 2(f(\hat{x}_0) - f^*)}{T\gamma D_1} +  \frac{D_2}{D_1} \left(\frac{N\sigma^2}{\sum_{l=1}^{N} M_l} + 2k\zeta^2\right) + \frac{ \gamma L \sigma^2}{(\sum_{l=1}^{N} M_l)D_1}.
\end{equation}
By setting $\gamma = \frac{1}{\sigma \sqrt{\frac{T}{\sum_{l=1}^{N} M_l}}}$ and $ T \geq \frac{48(\sum_{l=1}^{N} M_l)L^2k^2}{\sigma^2} $, we have
\begin{equation}\label{bound_corollary_basic}
16\gamma^2 L^2 k^2 = \frac{16\sum_{l=1}^{N} M_l}{\sigma^2T}L^2k^2 \leq \frac{1}{3}. 
\end{equation}
Now we can bound $D_1$ as
\begin{equation} \label{bound_D_1}
D_1 = 1-2kD_2 = 1 - \frac{16\gamma^2 L^2 k^2}{1-16\gamma^2 L^2 k^2} \geq \frac{1}{2}.
\end{equation}
Combining (\ref{bound_corollary_basic}) with $T \geq \frac{144(\sum_{l=1}^{N} M_l)^3}{\sigma^6}L^2k^2 \left( \frac{N\sigma^2}{\sum_{l=1}^{N} M_l} + 2k\zeta^2 \right)^2$, $D_2 \left(\frac{N\sigma^2}{\sum_{l=1}^{N} M_l} + 2k\zeta^2 \right)$ can be bounded as
\begin{eqnarray}\label{bound_D_2}
D_2 \left(\frac{N\sigma^2}{\sum_{l=1}^{N} M_l} + 2k\zeta^2 \right) &=& \frac{8\gamma^2 L^2 k}{1 - 16\gamma^2 L^2 k^2} \left(\frac{N\sigma^2}{\sum_{l=1}^{N} M_l} + 2k\zeta^2 \right)
\nonumber\\
&\leq& 12 \gamma^2 L^2 k \left(\frac{N\sigma^2}{\sum_{l=1}^{N} M_l} + 2k\zeta^2 \right)
\nonumber\\
&=& \frac{12\sum_{l=1}^{N} M_l}{\sigma^2T}L^2k \left(\frac{N\sigma^2}{\sum_{l=1}^{N} M_l} + 2k\zeta^2 \right)
\nonumber\\
&\leq& \frac{12\sum_{l=1}^{N} M_l}{\sigma^2\sqrt{T}}L^2k \left(\frac{N\sigma^2}{\sum_{l=1}^{N} M_l}+2k\zeta^2 \right) \cdot \frac{1}{\frac{12(\sum_{l=1}^{N} M_l)^{\frac{3}{2}}}{\sigma^3}Lk \left( \frac{N\sigma^2}{\sum_{l=1}^{N} M_l} + 2k\zeta^2 \right)}
\nonumber\\
&=& \frac{\sigma L}{\sqrt{T\sum_{l=1}^{N} M_l}}.
\end{eqnarray}
Substituting $\gamma = \frac{1}{\sigma \sqrt{\frac{T}{\sum_{l=1}^{N} M_l}}}$, (\ref{bound_D_1}) and (\ref{bound_D_2}) into (\ref{corollary_convergence}), we can get the final result:
\begin{equation}
\frac{1}{T} \sum_{t=0}^{T-1} \mathbb{E} \| \nabla f(\hat{x}_t) \|^2 \leq \frac{4\sigma\left(f(\hat{x}_0) - f^* + L\right)}{\sqrt{T\sum_{l=1}^{N} M_l}},
\end{equation}
which completes the proof.

\end{document}